\documentclass[letterpaper]{article} 
\usepackage{aaai25}  
\usepackage{times}  
\usepackage{helvet}  
\usepackage{courier}  
\usepackage{array}
\usepackage{tabularx}
\usepackage{longtable}
\usepackage{amssymb}
\usepackage[hyphens]{url}  
\usepackage{graphicx} 
\usepackage{natbib}  
\usepackage{siunitx}  
\usepackage{adjustbox} 
\usepackage{multirow}
\usepackage{booktabs}
\usepackage{caption} 
\usepackage{algorithm}
\usepackage{algorithmic}
\usepackage{pifont}

\usepackage{newfloat}
\usepackage{listings}
\DeclareCaptionStyle{ruled}{labelfont=normalfont,labelsep=colon,strut=off} 
\floatstyle{ruled}
\newfloat{listing}{tb}{lst}{}
\floatname{listing}{Listing}
\nocopyright 

\title{Rethinking Comprehensive Benchmark for Chart Understanding: A Perspective from Scientific Literature}

\author{
    Lingdong Shen\textsuperscript{\rm 1,2}$^{*}$,
    Qigqi\textsuperscript{\rm 1,2}\thanks{Equal contribution.},
    Kun Ding\textsuperscript{\rm 1,2}\thanks{Corresponding author.},
    Gaofeng Meng\textsuperscript{\rm 1,2},
    Shiming Xiang\textsuperscript{\rm 1,2}
}
\affiliations{
    \textsuperscript{\rm 1}School of Artificial Intelligence, University of Chinese Academy of Sciences\\
    \textsuperscript{\rm 2}MAIS, Institute of Automation, Chinese Academy of Sciences\\

    \{shenlingdong2022, qigeqi2023, kun.ding\}@ia.ac.cn\\     \{smxiang,gfmeng\}@nlpr.ia.ac.cn
}

\usepackage{bibentry}

\begin{document}

\maketitle

\begin{abstract}
Scientific Literature charts often contain complex visual elements, including multi-plot figures, flowcharts, structural diagrams and etc. Evaluating multimodal models using these authentic and intricate charts provides a more accurate assessment of their understanding abilities. However, existing benchmarks face limitations: a narrow range of chart types, overly simplistic template-based questions and visual elements, and inadequate evaluation methods. These shortcomings lead to inflated performance scores that fail to hold up when models encounter real-world scientific charts.
To address these challenges, we introduce a new benchmark, Scientific Chart QA (SCI-CQA), which emphasizes flowcharts as a critical yet often overlooked category. To overcome the limitations of chart variety and simplistic visual elements, we curated a dataset of 202,760 image-text pairs from 15 top-tier computer science conferences papers over the past decade. After rigorous filtering, we refined this to 37,607 high-quality charts with contextual information. SCI-CQA also introduces a novel evaluation framework inspired by human exams, encompassing 5,629 carefully curated questions, both objective and open-ended.
Additionally, we propose an efficient annotation pipeline that significantly reduces data annotation costs. Finally, we explore context-based chart understanding, highlighting the crucial role of contextual information in solving previously unanswerable questions. Our project will be released at \url{https://github.com/lydonShen/SCI-CQA}.
\end{abstract}

%

\section{Introduction}
Deep learning-based chart understanding focuses on tasks like chart question answering \cite{a1,dvqa,chartinstruct} and chart captioning \cite{a3,a4}. In recent years, the advent of multimodal models \cite{a5,a6} has revolutionized this field, as highlighted in \cite{pixels}. Foundational models have consistently set new performance records on chart understanding benchmarks. However, these improved results do not conclusively demonstrate that large models have achieved human-level capabilities in chart understanding.
\begin{figure*}[t]
\centering
\includegraphics[width=1\textwidth]{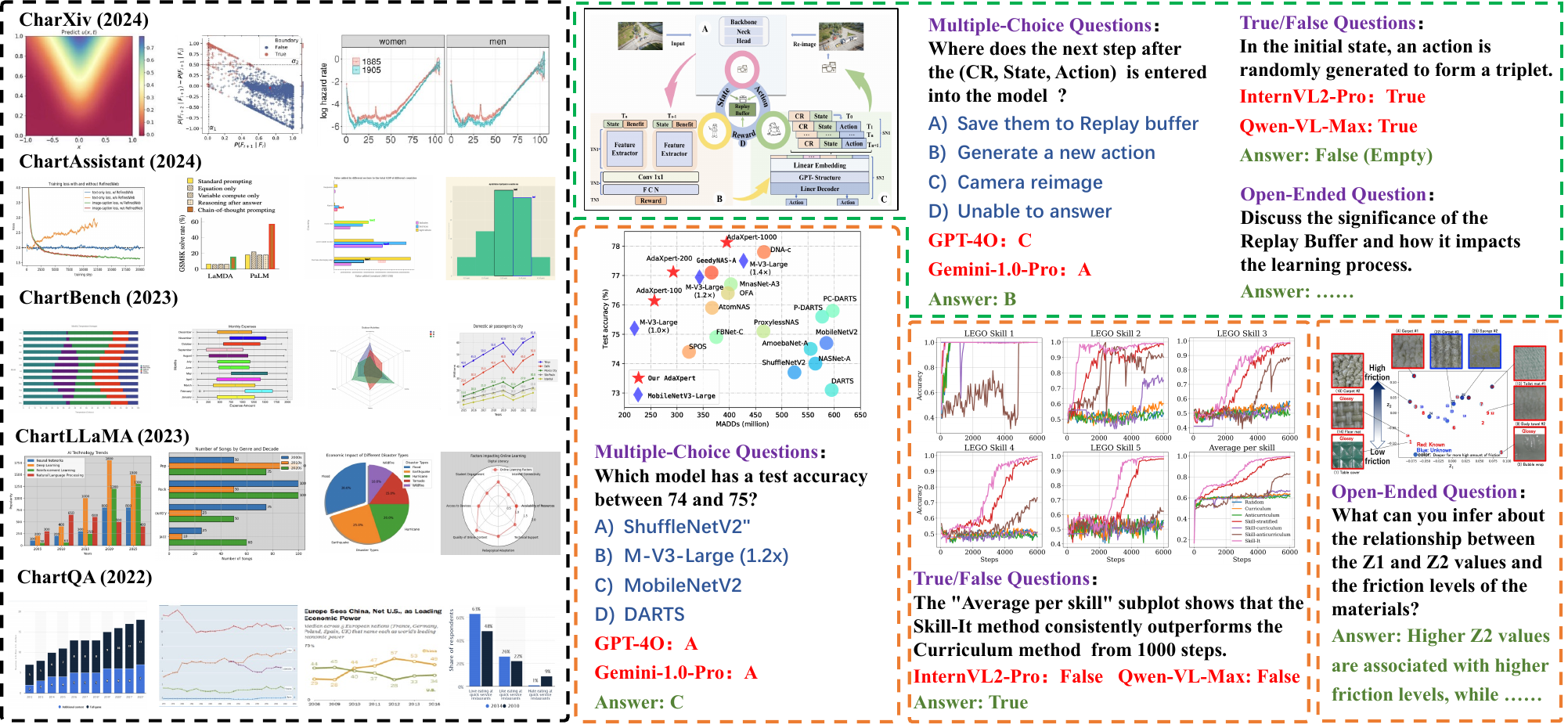} 
\caption{Comparison of SCI-CQA with previous work. The black box represents prior work, the green box highlights the SCI-CQA flowchart section, and the orange box represents the SCI-CQA data chart section. This comparison underscores SCI-CQA's advancements in data diversity and question complexity.}
\label{fig1}
\end{figure*}

Existing research encounters three critical issues. First, the charts used are limited in variety, characterized by fixed styles and simple visual representations, while overlooking challenging but essential types like flowcharts. For example, some studies rely on synthetic data to generate charts \cite{dvqa,figureqa}, or use web-sourced raw data with automated programs to create charts \cite{plotqa} (Left of Fig. \ref{fig1}). These generated charts have a significant complexity gap with scientific literature chart.

Second, the template-based questions and overly simplistic chart visual elements allows models to bypass the actual chart input, either by directly answering questions or by selecting correct answers based on hallucinations. For example, FigureQA \citep{figureqa} uses 15 templates to generate question-answer pairs, making it easy for large multimodal models \cite{a9,a6} to achieve high accuracy without truly understanding the chart.

Third, evaluation methods tend to be either only including subjective questions or objective questions, with insufficient differentiation between types of evaluation. This lack prevents a precise assessment of the models' specific strengths and weaknesses. For instance, \cite{chartbench} uses ACC+ for objective evaluation, while \cite{chartllama,charxiv} rely on GPT-4 scoring for open-ended assessment.

To ensure a more thorough and equitable assessment of models' chart understanding abilities, we have created the SCI-CQA benchmark, drawing from scientific literature. By utilizing the wealth of charts available in scientific papers, SCI-CQA addresses the issues identified in previous benchmarks.
We collected \textbf{202,760 image-text pairs} with context and captions from papers presented at 15 top-tier computer science conferences. Through a rigorous three-stage filtering process, we refined this dataset to \textbf{37,607 high-quality image-text pairs}. These pairs offer a diverse range of chart types and styles, rich in visual elements, varied in distribution, and include multiple sub-charts (see Right of Fig. \ref{fig1}). Notably, we are the first to collect and annotate scientific flowcharts and evaluate them as a distinct category. 

To avoid inaccurate performance evaluations due to template-based questions, we drew inspiration from \cite{charxiv} and proposed six types of base questions to broaden the scope of answers. By incorporating charts and textual context inputs, we generated questions that cannot be answered solely from the images. We then used GPT-4o to produce the corresponding question-answer pairs.
The question-answer pairs were then manually curated to form a high-quality validation set. As shown in Fig.\ref{fig1}, structure diagrams \cite{learnhowtosee} frequently cause leading multimodal models to make errors due to their complex visual elements and relationships. Additionally, real chart data—characterized by multiple subplots, image-chart combinations, and intricate data distributions—poses significant challenges to both open-source and proprietary models, resulting in suboptimal performance.

Finally, we introduced an evaluation method inspired by human exams, incorporating multiple-choice, true/false, and open-ended questions. We compiled a question bank of \textbf{5,629 questions}, all of which were reviewed by master's and doctoral students specializing in computer science. Given that these data were sourced from the computer science domain, the review by highly educated professionals ensured both the diversity and accuracy of the questions.

Additionally, we developed an \textbf{automated annotation pipeline} to cut the costs associated with proprietary model annotation. Using the contextual data provided by SCI-CQA, we also explored whether complex reasoning questions, which cannot be answered solely from images, can be effectively resolved with the assistance of contextual information. In summary, our contributions include rethinking a more comprehensive evaluation method for chart understanding from the perspective of scientific literature, thereby promoting advancements in the field of chart understanding.

\section{Related Work}
\subsection{Chart understanding benchmark}
Previous work reveals room for improvement in data diversity, question complexity, and evaluation fairness.

In terms of data and visual element diversity, previous studies such as \cite{dvqa, figureqa, plotqa, flowvqa} rely on charts synthesized from real data, resulting in limited data diversity and simple visual elements. These charts often lack the complexity and variety found in actual scientific literature, which typically features a wide range of presentation styles, complex data distributions, and rich visual information (see the right side of Fig. \ref{fig1}). While \cite{chartllama} utilizes multimodal large models to generate code for creating charts, their approach includes only 4-8 chart types. \cite{chartx} expanded the chart types to 18 categories but still faced challenges with visual element diversity. Additionally, \cite{charxiv} used real chart data to address these limitations but overlooked flowcharts, an essential chart type. 

SCI-CQA, inspired by \cite{charxiv}, utilizes natural data charts for model evaluation and includes flowcharts (such as model architecture diagrams and algorithm flowcharts), which are rich in information and crucial to scientific literature. Natural data charts and flowcharts facilitate the integration of chart understanding into the context of scientific literature comprehension.

\begin{figure*}[t]
\centering
\includegraphics[width=1\textwidth]{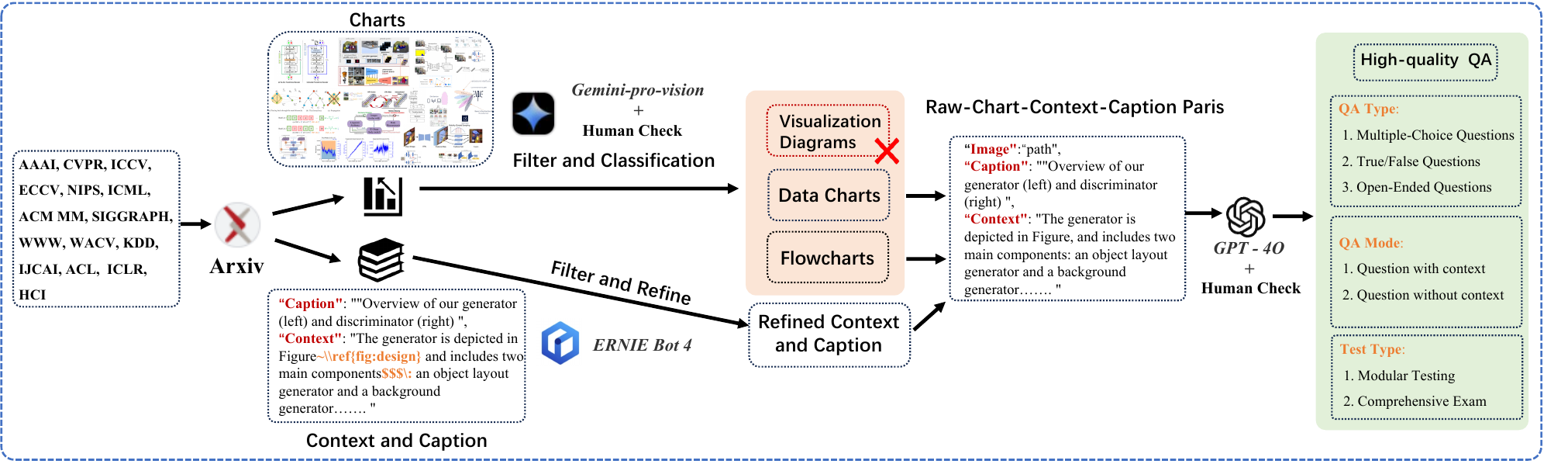} 
\caption{SCI-CQA data processing pipeline, multiple proprietary models is employed to monitor and enhance data quality, supplemented by rigorous manual verification at key stages, whichs led to the creation of a high-quality evaluation dataset.}
\label{fig2}
\end{figure*}
Regarding the scientific rigor of question-answering pairs, previous works such as \cite{dvqa} used template-based questions that are relatively simple and limited in variety. \cite{chartbench} defined five question types from perception and probability perspectives, while \cite{charxiv} created templates for 19 questions across five task types, all answerable solely from the image. \cite{mathvista} focused on mathematical reasoning in visual contexts, and \cite{mapqa} concentrated on map understanding. In contrast, SCI-CQA builds on \cite{charxiv} by introducing a sixth question type: scientific reasoning. It also incorporates chart context to propose questions that cannot be answered by the chart alone. This approach ensures that when models encounter reasoning questions that cannot be solved with visual input alone, they must acknowledge their inability to answer, thus mitigating hallucination issues. As illustrated in Fig.\ref{fig1}, SCI-CQA's questions are more challenging, posing difficulties even for state-of-the-art proprietary models.

Regarding evaluation methods, \cite{chartbench} utilizes an objective ACC+ mode, providing more standardized results. In contrast, \cite{charxiv} and \cite{mmc} employ a free-form question-answering mode, which assesses the model's overall capability in a more flexible manner. SCI-CQA integrates both approaches and adopts a human exam-inspired framework. It categorizes questions into multiple-choice, true/false, and open-ended formats, offering a more comprehensive evaluation of a model's abilities.

Some works \cite{arxivqa, scigraphqa, RealCQA} have made contributions by addressing the need for datasets in the field of scientific literature chart understanding. However, the absence of manual review limits the accuracy of model performance assessments. Our review reveals that only about 47\% of the QA pairs generated by GPT-4o were correct. Furthermore, the datasets they used lack contextual information, which hinders the model's ability to handle more complex tasks. On the other hand, \cite{multimodalimage} and \cite{flowvqa} have recognized flowcharts as a distinct chart modality. However, the synthetic flowcharts in their studies differ in complexity and diversity from those in SCI-CQA.

\section{SCI-CQA}
\subsection{Data processing pipeline }
\noindent \textbf{[1].Data source.} 
The dataset sources include 15 top-tier conferences (shown in Fig.\ref{fig2}) covering computer vision, natural language processing, machine learning, multimedia information processing, and emerging interdisciplinary fields. To ensure high data quality, SCI-CQA leverages the expertise of graduate students and PhD candidates in computer science for meticulous data filtering and question-answer reviews. We obtained LaTeX source files from these conferences over the past decade, totaling 75,506 articles. After filtering, we retained 54,557 papers, excluding those with unavailable source files or insufficient chart data.

\begin{figure*}[t]
\centering
\includegraphics[width=1\textwidth]{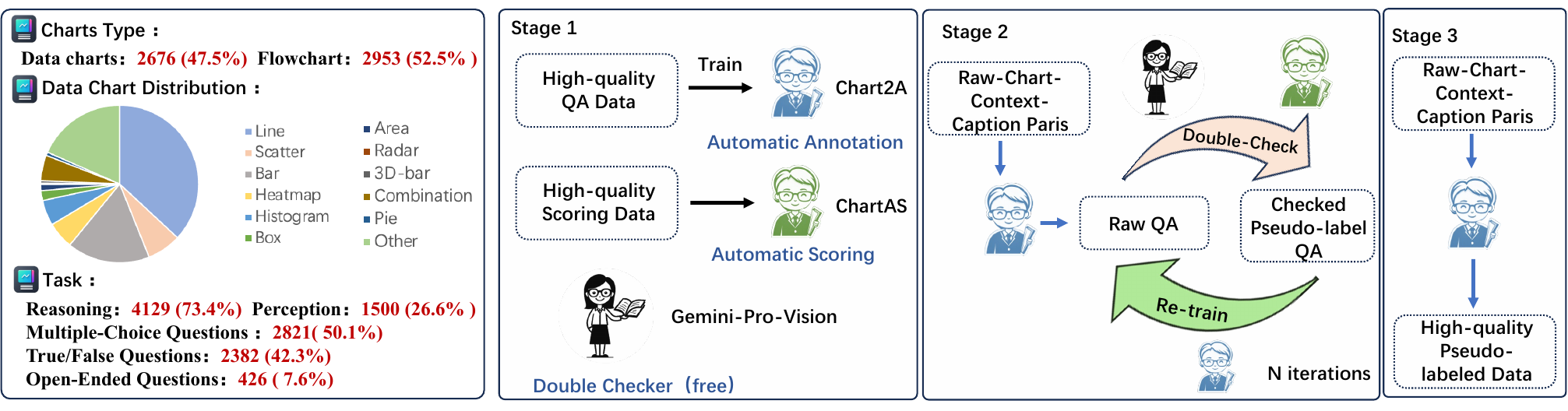} 
\caption{Data statistics (left) and automated annotation tool pipeline (right). High-quality annotated data and scoring data from model evaluations are used to train the open-source model, which iteratively generates additional pseudo-labeled data.}
\label{fig3}
\end{figure*}
\noindent \textbf{[2].Raw chart-context-caption pairs.} 
From the perspective of scientific literature comprehension, interpreting charts involves more than understanding visual elements, it requires addressing complex reasoning tasks by integrating information from both the chart and its contextual surroundings. Unlike previous work, our data collection method uses a triplet representation: (chart, caption, context).

Charts are classified using the proprietary Gemini model \cite{gemini}, which distinguishes between data charts, flowcharts, and visualization analysis charts. We exclude the visualization analysis chart(different tasks can vary greatly and are difficult to assess uniformly) and manually review the data charts and flowcharts. The "caption" consists of the textual description of the chart, extracted from the \texttt{caption} tags in LaTeX source files. The "context" includes relevant surrounding text, cited via \texttt{ref}. Given that the extracted text often contains special symbols and LaTeX commands, we refined the text descriptions using techniques from \cite{ernie} to ensure high-quality textual data.

\noindent \textbf{[3].High-quality QA pairs.} 
As illustrated in the left half of the figure, the number of flowcharts and data charts in SCI-CQA is nearly balanced. SCI-CQA is the first to compile and evaluate a comprehensive multimodal QA dataset for natural flowcharts, which encompass widely used scientific illustrations such as model structure diagrams, algorithm flowcharts, and model training pipelines. Additionally, SCI-CQA features 11 common types of data charts, including line charts, scatter plots, and pie charts. 
SCI-CQA also includes charts with multiple subplots, combinations of images and data visualizations, and hybrid charts that integrate two or more visualization types. Notably, it features specialized charts such as forest plots, correlation matrix charts, Gantt charts, violin plots, density plots, and Pareto charts, which are often overlooked in previous work but are crucial in scientific literature.

Regarding QA types, the dataset is divided into perception and reasoning tasks. While perception tasks have received significant attention in previous work, SCI-CQA primarily focuses on reasoning tasks. The question types maintain a \textbf{5:4:1} ratio of multiple-choice, true/false, and open-ended questions. All questions are available in the question bank, allowing for custom creation of exams.
\subsection{Human-Inspired evaluation method}
Evaluating models using a combination of multiple question types (such as multiple-choice, true/false, and open-ended questions) provides a more comprehensive and realistic assessment of their understanding, reasoning, and generative abilities, making it more accurate than relying on a single question format. This method includes 4 types questions. Multiple-Choice Questions: The model selects the single correct answer from four options. True/False Questions: The model determines whether a given statement is correct or incorrect. Open-Ended Questions: The model provides a free-form response based on the prompt. Unanswerable Questions: The model must indicate if it cannot provide an answer based on the question and input.

We created a question bank with 5,629 QA pairs, allowing for flexible test construction and scoring. As part of our evaluation set, we designed three test papers, each comprising: 40 multiple-choice questions, 40 true/false questions, 4 open-ended questions. Multiple-choice and true/false questions are valued at 1 point each, while open-ended questions are valued at 5 points, resulting in a maximum possible score of 100 points. For objective questions, we directly assess the correctness of answers. For open-ended questions, we use GPT-4o-based scoring (see the appendix for details).
\subsection{Automated labeling tools}
Data annotation is highly resource-intensive and costly, especially when using proprietary models for generating question-answer pairs. This study explores fine-tuning an open-source model to reduce annotation costs effectively. 
The goal is to utilize the trained model for data annotation, enabling the creation of a lager training dataset.

\noindent \textbf{[1].Chart2A and ChartAS.}
We used LLaVA \cite{llava} as the base model and fine-tuned it with high-quality annotated QA pairs while preserving the original model structure and training parameters. The fine-tuning dataset included 45k GPT-4o-annotated QA pairs, which had not undergone manual verification. For this purpose, we utilized all GPT-4o-based scores collected during testing, totaling 30k scores. This fine-tuning process was completed in the first stage, as shown in Fig.\ref{fig2}-Stage1. 
In this training process, we employed a two-epoch approach for training LLaVA. In the first epoch, the Large Language Model (LLM) and Visual Encoder were fixed while training the projector. In the second epoch, we used LoRA to fine-tune the LLM and SFT to fine-tune the projector. For subsequent training with mixed datasets, only one epoch training was used (LoRA training). 
After the completion of the first stage, we obtained two key models: \textbf{Chart2A}, used for data annotation, and \textbf{ChartAS}, used for question-answer scoring.

\noindent \textbf{[2].Automated labeling pipeline.}
After completing the initial training stage, we employed the free proprietary model Gemini-Pro-Vision and the fine-tuned ChartAS as supervisors to label 37k filtered chart-caption-context datasets. This labeling process generated 350k pseudo-labeled data points. These pseudo-labels were then verified and scored by both models, resulting in a validated set of pseudo-labeled data. This dataset was subsequently integrated into Chart2A's training set for further retraining (one epoch retraining). This iterative process was repeated N times, culminating in the production of the final pseudo-labeled dataset. Throughout this procedure, no paid proprietary models were used, achieving significant cost savings.

\begin{table*}[t]
\centering
\renewcommand{\arraystretch}{1.2}  
\begin{adjustbox}{max width=\textwidth}
\begin{tabular}{lcccccccccc}
\toprule\toprule
\textbf{Dataset} & \textbf{RD} & \textbf{RC} & \textbf{TQ} & \textbf{FTQ} & \textbf{OVA} & \textbf{IF} & \textbf{ET} & \textbf{HC} & \textbf{CT} & \textbf{Eval.} \\ 
\midrule
FigureQA      & \ding{55}  & \ding{55}  & \ding{51}  & \ding{55}  & \ding{55}  & \ding{55}  & \ding{55}  & \ding{55}  & 3  & 1.5k  \\ 
DVQA          & \ding{55}  & \ding{55}  & \ding{51}  & \ding{55}  & \ding{51}  & \ding{55}  & \ding{55}  & \ding{55}  & 3  & 580k  \\ 
PlotQA        & \ding{51}  & \ding{55}  & \ding{51}  & \ding{55}  & \ding{51}  & \ding{55}  & \ding{55}  & \ding{55}  & 3  & 33.7k \\ 
ChartQA       & \ding{51}  & \ding{51}  & \ding{55}  & \ding{51}  & \ding{51}  & \ding{55}  & \ding{55}  & \ding{51}  & 3  & 1.5k  \\ 
ChartBench    & \ding{51}  & \ding{51}  & \ding{55}  & \ding{51}  & \ding{55}  & \ding{55}  & \ding{55}  & \ding{51}  & 4  & 2k    \\ 
MMC           & \ding{51}  & \ding{51}  & \ding{55}  & \ding{51}  & \ding{51}  & \ding{55}  & \ding{55}  & \ding{51}  & 9  & 2.1k  \\ 
ChartX        & \ding{51}  & \ding{51}  & \ding{55}  & \ding{51}  & \ding{51}  & \ding{55}  & \ding{55}  & \ding{51}  & 18 & 6k    \\ 
CharXiv       & \ding{51}  & \ding{51}  & \ding{51}  & \ding{51}  & \ding{51}  & \ding{55}  & \ding{55}  & \ding{51}  & 19 & 2.3k  \\ 
\toprule
\textbf{Ours}        & \ding{51}  & \ding{51}  & \ding{51}  & \ding{51}  & \ding{51}  & \ding{51}  & \ding{51}  & \ding{51}  & \textbf{21} & \textbf{5.6k} \\ 
\bottomrule
\bottomrule
\end{tabular}
\end{adjustbox}
\caption{Comparison of datasets across multiple evaluation aspects. The abbreviations used in the table are: RD (Real Data), RC (Real Chart), TQ (Template Question), FTQ (Free-Form Question), OVA (Open Vocab Answer), IF (Include Flowchart), ET (Exam Test), HC (Human Check), CT (Chart Type), Eval. (Number of Evaluation Samples).}
\label{tab:dataset_comparison}
\end{table*}
\subsection{Analysis of the Comparison Across Datasets}

From the comparison (As show in Tab.1), we can observe several key difference:

\noindent \textbf{[1]. Diversity in Question Types and Answer Formats:} SCI-CQA include a broader range of features such as free-form questions and open vocabulary answers, which allow for more flexible and complex interactions with the model. In contrast, datasets like FigureQA and DVQA focus primarily on template questions, with no open-form answer options, which can limit the scope of evaluation.

\noindent \textbf{[2]. Inclusion of Flowcharts}: The Include Flowchart feature is notably present in the SCI-CQA dataset, where all evaluation aspects incorporate flowchart-based visual understanding. This feature is crucial for tasks that require structured reasoning, highlighting SCI-CQA as a more complex and holistic benchmark compared to datasets like Charxiv and ChartX, which exclude flowchart integration.

\noindent \textbf{[3]. Evaluation Samples and Human-check}: SCI-CQA stands out with the highest number of evaluation samples (5.6k), along with a broad scope of evaluation criteria, including multiple task types (perception, reasoning) and comprehensive dataset features. This large scale human-check benchmark allows for a more robust performance analysis and benchmarking. 

\noindent \textbf{[4]. Real Data and Chart Types}: Real chart inclusion helps enhance the realism and relevance of model training and evaluation, making SCI-CQA a critical component for evaluating chart-based question answering.It can also be noted that SCI-Cqa contains the richest variety of charts. The combination of these two strengths makes SCI-CQA the most comprehensive benchmark for chart understanding.

\section{Experiment}
\subsection{Experiment Setting}
We evaluated 14 models on SCI-CQA, including 11 open-source multimodal models and 3 proprietary models: GPT-4o, GPT-4-turbo, and GPT-4o-mini \cite{a9}. The open-source models tested were:Chartllama \cite{chartllama}, SPHINX \cite{lin2023sphinx}, CogVLM \cite{cogvlm}, DocOwl 1.5 \cite{mplug}, Internlm-xcomposer-V2 \cite{Internlm-xcomposer2}, InternVL2-8B \cite{internvl}, LLaVA-V1.6-13B \cite{llava-next}, Mini-Gemini \cite{mini}, mPLUG-OwI2\cite{ye2024mplug}, Qwen-VL-Chat \cite{bai2023qwen}, Cogagent \cite{cogagent}. In all experiments, the models were tested with their original input-output settings, parameters, and weight configurations. Separate tests were conducted for flowcharts and data charts. Training tasks were performed on 4 NVIDIA A100 GPUs.

Multiple-choice and true/false questions were scored objectively, while open-ended questions were evaluated using a GPT-4o-based scoring strategy. 
For multiple-choice questions, the model must select the correct answer from four options. For true/false questions, the model must provide the correct judgment. The accuracy of these responses is calculated and scaled to a maximum score of 100. For open-ended questions, the scoring range is from 0 to 5 points, with the scoring criteria detailed in the appendix.
The Composite score was calculated by scaling the open-ended question scores to a 0-100 range and averaging them with the scores from Avg MC) and Avg TF.
\begin{table*}[t]
\centering
\renewcommand{\arraystretch}{1}
\begin{adjustbox}{max width=\textwidth}
\begin{tabular}{lccccccccc}
\toprule
\toprule
\multirow{3}{*}{Type} & \multirow{3}{*}{Model} & \multicolumn{5}{c}{Task Performance} & \multirow{3}{*}{Avg MC} & \multirow{3}{*}{Avg TF} & \multirow{3}{*}{Composite Score} \\ \cmidrule(lr){3-7}
 & & \multicolumn{2}{c}{Perception Task} & \multicolumn{2}{c}{Reasoning Task} & \multirow{2}{*}{Open} & & & \\ \cmidrule(lr){3-4} \cmidrule(lr){5-6}
 & & MC & T/F & MC & T/F & & & & \\ \midrule
\multirow{11}{*}{Open} 
 & Chartllama          & 14.79 & 17.50 & 12.61 & 17.22 & 1.65 & 13.06 & 17.29 & 21.12 \\ 
 & SPHINX              & 17.18 & 12.81 & 16.72 & 13.28 & 1.94 & 17.08 & 13.16 & 23.01 \\ 
 & LLaVA-V1.6-13B      & 25.08 & \textbf{70.63} & 25.15 & 68.86 & 2.55 & 25.13 & 68.31 & 48.15 \\ 
 & Qwen-VL-Chart       & 23.79 & 45.62 & 28.46 & 55.47 & 2.51 & 27.51 & 52.97 & 43.56 \\ 
 & mPLUG-Owl2          & 34.41 & 28.12 & 28.63 & 27.10 & 2.01 & 29.82 & 27.36 & 32.46 \\ 
 & Mini-Gemini         & \textbf{41.16} & 9.69 & 34.19 & 13.18 & 2.60 & 35.62 & 12.29 & 33.30 \\ 
 & DocOwl1.5           & 39.23 & 24.69 & 37.68 & 23.06 & 1.23 & 37.99 & 23.47 & 28.67 \\ 
 & Cogagent-vqa        & 32.48 & 15.00 & 43.15 & 14.45 & 2.64 & 40.96 & 14.59 & 36.12 \\ 
 & Cogvlm              & 30.55 & 22.81 & 45.23 & 31.67 & 2.71 & 42.22 & 29.42 & 41.95 \\ 
 & InternLM-XcompserV2 & 32.15 & 36.88 & 46.39 & 34.54 & 2.76 & 43.47 & 35.13 & 44.60 \\ 
 & InternVL2           & 40.19 & 66.25 & \textbf{50.62} & \textbf{71.94} &\textbf{3.14} & \textbf{48.48} & \textbf{70.50}& \textbf{60.59} \\ 
\midrule
\multicolumn{2}{c}{\textbf{Average for Open Models}} & \multicolumn{4}{c}{}& \textbf{2.34} & \textbf{32.85} & \textbf{33.14} & \textbf{37.60} \\ 
\midrule
\multirow{3}{*}{Private} 
 & GPT-4o-mini  & 35.05 & 70.63 & 47.39 & 70.24 & 3.72 & 44.85 & 70.34 & 63.20 \\ 
 & GPT-4-turbo  & \textbf{50.16} & 73.75 & \textbf{58.67} & 75.13 & 3.91 & \textbf{56.93} & 74.78 & 69.97 \\ 
 & GPT-4o       & 41.48 & \textbf{78.75} & 53.44 & \textbf{77.47} & \textbf{4.45} & 50.99 &\textbf{77.80} & \textbf{72.60} \\ 
\midrule
\multicolumn{2}{c}{\textbf{Average for Private Models}} & \multicolumn{4}{c}{}& \textbf{4.03} & \textbf{50.92} & \textbf{74.31} & \textbf{68.61} \\ 
\bottomrule
\bottomrule
\end{tabular}
\end{adjustbox}
\caption{Comparison of flowchart understanding performance across 14 models, including Perception and Reasoning tasks with multiple-choice (MC), true/false (TF), and open-ended questions. Models categorized as Open-Source or Proprietary.}
\label{tab:performance_comparison}
\end{table*}
\subsection{Performance analysis of flowchart}
\noindent \textbf{[1].All the models performed poorly on flowchart understanding.}
Analysis of Tab.2 reveals that both open-source and proprietary models have substantial room for improvement in flowchart understanding. For instance, even the latest InternVL2 achieved only a \textbf{60.59} composite score. Among open-source models, Chartllama scored just \textbf{21.12}, lower than its \textbf{69.66} average performance on the ChartQA dataset. This discrepancy highlights that existing benchmarks often overestimate the chart understanding capabilities of multimodal models. Furthermore, it underscores the need for increased focus on flowcharts as a crucial chart type. The purely natural flowcharts in SCI-CQA represent a challenging task for multimodal models, emphasizing the necessity for more dedicated research in this area.In T/F perception tasks, Mini-Gemini scored only 9.69 points due to its poor instruction-following ability, often failing to produce "True" or "False" answers. As a result, any incorrect output is counted as an error.

\noindent \textbf{[2].Performance Gap Between Open-Source and Proprietary Models.} Based on the average performance data from Tab.2, notable differences between open-source and proprietary models are evident:

\textbf{Open-Ended Questions}: Proprietary models significantly outperform open-source models, with GPT-4o achieving a score of 4.45 compared to DocOwl1.5's 1.23, resulting in a difference of 3.22. 

\textbf{Objective Questions}: Proprietary models also lead in this category. They average a score of 50.92 in multiple-choice questions versus 32.85 for open-source models, showing a difference of 18.07. The disparity is even greater in true/false questions, where proprietary models exceed open-source models by 41.17.

These results highlight a clear advantage for proprietary models, particularly in handling open-ended questions. However, the performance gap on objective question is gradually narrowing. For example, InternVL2's overall score is only 2.61 points behind GPT-4o-mini. InternVL2 has also surpassed GPT-4o-mini in multiple-choice and true/false questions, with improvements of 3.63 and 0.16, respectively. These advancements are attributed to effective feature alignment, retraining of the visual encoder, and higher input resolution. Despite these improvements, InternVL2 still trails GPT-4o by 12.01 in the composite score.

\noindent \textbf{[3].Reasoning Performance Highlights Model Differences Once Perception Abilities Reach a Threshold.} Despite a positive correlation between perception and reasoning, superior performance in perception does not guarantee better reasoning abilities. For example, while mPLUG-Owl2 achieves a perception score of 34.41, its reasoning score is only 28.63. In contrast, InternLM-XcomposerV2 scores 32.15 in perception and 46.39 in reasoning. This means that mPLUG-Owl2 excels in perception by 2.26 points but lags 17.76 points behind in reasoning.
This discrepancy underscores that, although reasoning tasks are somewhat dependent on perception, they are not entirely contingent upon them. Perception and reasoning are partially independent aspects of model performance. A model with strong perception capabilities does not necessarily excel in reasoning, but a model with weak perception abilities is likely to struggle with reasoning tasks. For instance, SPHINX, with a perception score of only 17.18, also performs poorly in reasoning.
When perception scores exceed 30, reasoning performance tends to exhibit greater independence. Thus, while strong perception capabilities provide a necessary foundation, reasoning abilities can be evaluated more independently as a distinct aspect of model performance once a certain threshold of perception is reached.
\begin{table*}[t]
\centering
\renewcommand{\arraystretch}{1}
\begin{adjustbox}{max width=\textwidth}
\begin{tabular}{lccccccccc}
\toprule
\toprule
\multirow{3}{*}{Type} & \multirow{3}{*}{Model} & \multicolumn{5}{c}{Task Performance} & \multirow{3}{*}{Avg MC} & \multirow{3}{*}{Avg TF} & \multirow{3}{*}{Composite Score} \\ \cmidrule(lr){3-7}
 & & \multicolumn{2}{c}{Perception Task} & \multicolumn{2}{c}{Reasoning Task} & \multirow{2}{*}{Open} & & & \\ \cmidrule(lr){3-4} \cmidrule(lr){5-6}
 & & MC & T/F & MC & T/F & & & & \\ \midrule
\multirow{11}{*}{Open} 
 & Chartllama          & 18.60 & 47.98 & 20.05 & 52.84 & 1.60 & 19.46 & 49.51 & 33.66 \\ 
 & LLaVA-V1.6-13B      & 25.00 & 40.91 & 26.65 & 42.13 & \textbf{2.33} & 25.98 & 41.75 & 38.11\\ 
 & Qwen-VL-Chart       & 31.59 & 46.31 & 27.28 & 42.00 & 2.00 & 28.97 & 48.35 & 39.11 \\ 
 & mPLUG-Owl2          & 23.64 & 50.00 & 26.65 & 54.48 & 1.55 & 25.44 & 53.08 & 36.51 \\ 
 & Mini-Gemini         & 43.99 & 43.75 & 26.14 & \textbf{55.27} & 2.19 & 33.18 & 49.24 & 42.07 \\ 
 & DocOwl1.5           & 33.91 & 56.53 & 26.14 & 51.50 & 1.25 & 29.20 & 53.08 & 35.76 \\ 
 & Cogagent-vqa        & 42.25 & 55.97 & 32.23 & 54.62 & 2.07 & 36.17 & \textbf{55.04} & 44.20 \\ 
 & Cogvlm              & 44.19 & 52.80 & 35.15 & 50.00 & 2.32 & 38.70 & 51.92 & 45.67 \\ 
 & InternLM-XcompserV2 & \textbf{55.62} & \textbf{57.96} & \textbf{47.84} & 52.93 & 2.12 & \textbf{50.88} & 54.42 & \textbf{49.23} \\ 
\midrule
\multicolumn{2}{c}{\textbf{Average for Open Models}} & \multicolumn{4}{c}{}& \textbf{1.94} & \textbf{32.00} & \textbf{50.71} & \textbf{40.50} \\ 
\midrule
\multirow{3}{*}{Private} 
 & GPT-4o-mini  & 55.04 & 61.36 & 41.88 & 53.97 & 3.03 & 47.05 & 56.29 & 54.65 \\ 
 & GPT-4-turbo  & \textbf{62.21} & 71.31 & 41.12 & 63.72 & 3.50 & 49.43 & 66.10 & 61.84 \\ 
 & GPT-4o       & 59.30 & \textbf{75.85} & \textbf{43.91} & \textbf{68.27} & \textbf{3.89} &\textbf{49.96} & \textbf{70.74} &\textbf{66.17} \\ 
\midrule
\multicolumn{2}{c}{\textbf{Average for Private Models}} & \multicolumn{4}{c}{}& \textbf{3.47} & \textbf{48.81} & \textbf{64.38} & \textbf{60.86} \\ 
\bottomrule
\bottomrule
\end{tabular}
\end{adjustbox}
\caption{Comparison of data chart understanding performance across 12 models.}
\label{tab:performance_comparison}
\end{table*}
\subsection{Performance analysis of data chart}
\noindent \textbf{[1].The Performance Gap Between Open-Source and Private Models Narrows on data charts.} We tested 12 models using data charts, revealing a narrowing performance gap between open-source and proprietary models. For example, InternLM-XcomposerV2 outperformed GPT-4o-mini in perception tasks for multiple-choice questions (55.62 vs. 55.04). It also surpassed GPT-4o in reasoning tasks for multiple-choice questions (47.84 vs. 43.91), with an overall average score advantage of 0.92 points. This suggests that open-source models are gaining a competitive edge in traditional chart interpretation.

Mini-Gemini exceeded GPT-4o-mini in reasoning tasks for true/false questions (55.27 vs. 53.97), although it still trails GPT-4o by 13 points. LLaVA-V1.6-13B also demonstrated significant progress in open-ended questions, scoring 2.33 and closing the gap with proprietary models.
These positive results can largely be attributed to the incorporation of chart data during the pre-training phase. Models like InternLM-XcomposerV2 and LLaVA-V1.6-13B, which included chart data in their training, showed improvements in performance. This underscores the importance of including diverse chart types, such as flowcharts, in benchmark to ensure a comprehensive assessment of multimodal models.

For instance, while InternLM-XcomposerV2 achieved state-of-the-art results with data charts, its average score on multiple-choice questions involving flowcharts was lower compared to data charts (43.47 vs. 50.88). This discrepancy highlights the critical role of diverse chart types in accurately evaluating a model’s capabilities.
\begin{figure*}[t]
\centering
\includegraphics[width=1\textwidth]{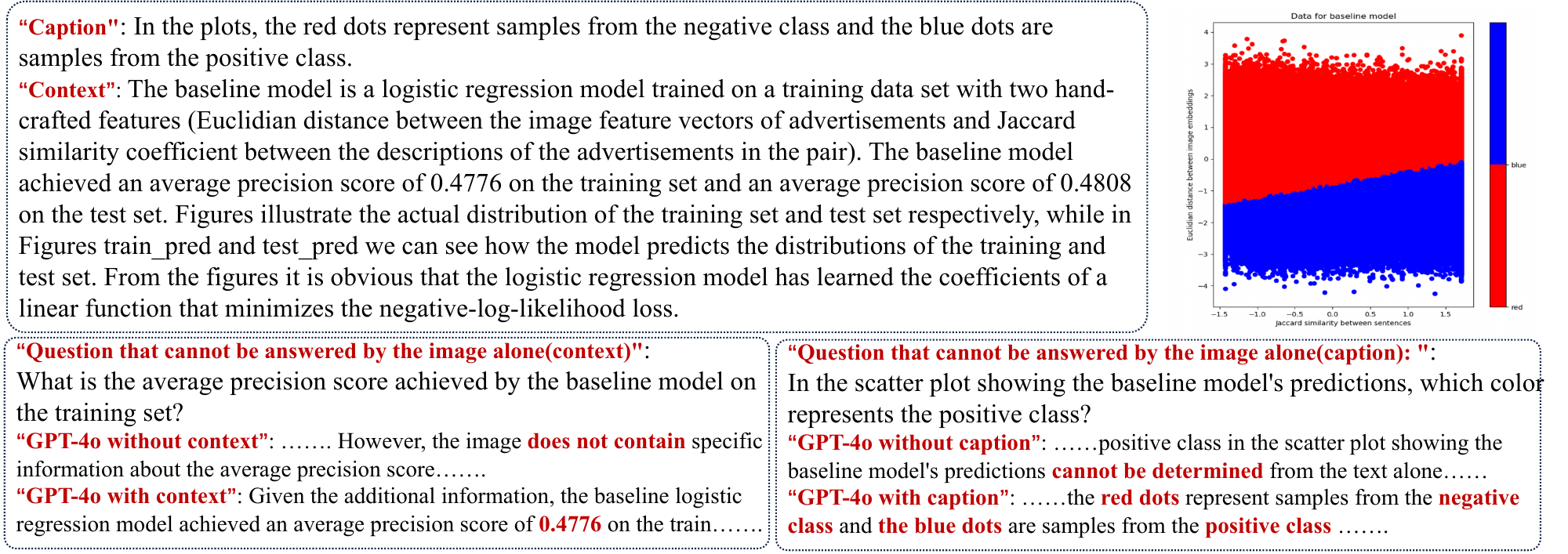} 
\caption{Impact of refined context and captions on answering previously unanswerable questions.}
\label{fig4}
\end{figure*}

\noindent \textbf{[2]. A Single Question Type Can't Fully Distinguish Model Performance}
When results for a specific type of question are clustered within a narrow range, it becomes challenging to differentiate model performance. For example, in Tab.3, the average scores for open-source models on multiple-choice are mostly between 49 and 54, making it difficult to discern performance differences. In contrast, a composite scoring approach broadens the performance distribution to between 33 and 49, offering a clearer view of performance variations.
For instance, the gap between mPLUG-Owl2 and Cogagent-vqa in multiple-choice questions is only 1.96 points, but this difference widens to 10.73 points when using the composite score. This illustrates that employing diverse evaluation methods can provide a more effective assessment of a model’s true capabilities.

\subsection{Effectiveness of Automated Annotation Pipeline}
Tab.4 illustrates the effectiveness of fine-tuning open-source models using data annotated by an automated pipeline. Specifically, models fine-tuned with Chart2A data, which underwent five iterations of annotation, show notable performance improvements compared to direct model testing (11.80 vs. 17.93 on data chart multiple-choice questions). This underscores the value of pseudo-labeled data in enhancing model performance.
However, the improvement in performance for open-ended questions on data charts is relatively modest (1.01 vs. 0.98 vs. 1.17), while for flowcharts, there is a more promising trend (1.38 vs. 1.51 vs. 1.53).
\begin{table}[h!]
\centering
\renewcommand\arraystretch{1.2}
\resizebox{\columnwidth}{!}{
\begin{tabular}{c|l|l l l l}
\toprule
\toprule
Chart Type& Data Setting & MC & TF &Open & Score\\ 
\midrule
\multirow{4}{*}{Data} & W/O Chart2A      & 11.80  & 31.13  & 1.01 & 21.04\\
                               & Chart2A / 1 It   & 12.87  & 36.75  & 0.98&23.07 \\
                               & Chart2A / 5 It   & 17.93  & 44.42  & 1.17&28.58 \\
\midrule
\multirow{4}{*}{Flow} & W/O Chart2A      & 11.35  & 13.32  & 1.38 &17.42\\
                               & Chart2A / 1 It   & 12.47  & 15.94  & 1.51&19.54 \\
                               & Chart2A / 5 It   & 13.92  & 18.64  & 1.53 & 21.05\\
\bottomrule
\bottomrule
\end{tabular}}
\caption{Performance comparison of fine-tuning LLaVA with data annotated by the automated annotation pipeline}
\end{table}

Two main factors contribute to these results. \textbf{Proportion of Open-Ended Questions}: The training dataset contains fewer open-ended questions compared to MC and TF, with a ratio of approximately 2:2:1. This imbalance affects the overall effectiveness of fine-tuning on open-ended questions.\textbf{Quality of Generated Questions}: The generated open-ended questions are relatively low in quality, averaging around 36 words in length, compared to the more detailed data produced by GPT-4o. This issue arises from the base model's design. LLaVA, while employing a simple and effective multimodal structure, lacks specific adaptations for generating high-quality open-ended questions.

Overall, Chart2A demonstrates effectiveness, with improvements seen in both objective and subjective questions. Future enhancements could involve using a more powerful base model, such as InternVL2, for fine-tuning or increasing the number of iterations to further enhance data quality.
\begin{table}[h!]
\centering
\renewcommand\arraystretch{1.2}
\resizebox{\columnwidth}{!}{
\begin{tabular}{cccccc}
\toprule
\toprule
Type& Model & MC & TF & Open & Score \\ 
\midrule
\multirow{4}{*}{Data}       & Chartllama.  & 8/40   & 18/40  & 5/20  & 31/100   \\
                            & Cogvlm       & 14/40  & 18/40  & 9/20  & 41/100   \\
                            & Qwen-VL-chat & 9/40   & 18/40  & 11/20 & 38/100    \\
                            & Human        & 36/40  & 38/40  & 17/20 & 91/100  \\
\midrule
\multirow{4}{*}{Flow}       & Chartllama.  &  8/40  & 12/40  & 7/20  & 27/100  \\
                            & Cogvlm       & 13/40  & 5/40   & 13/20 & 31/100  \\
                            & Qwen-VL-chat &  6/40  & 14/40  & 12/20 & 32/100  \\
                            & Human        & 39/40  & 35/40  & 16/20 & 90/100  \\
\midrule
\multirow{4}{*}{Data+Flow}  & Chartllama.  & 2/40  & 14/40   & 4/20  & 20/100   \\
                            & Cogvlm       & 6/40  & 17/40   & 5/20  & 28/100     \\
                            & Qwen-VL-chat & 9/40  & 17/40   & 11/20 & 37/100     \\
                            & Human        & 37/40 & 35/40   & 16/20 & 88/100     \\
\bottomrule
\bottomrule
\end{tabular}}
\caption{Performance comparison of new evaluation method}
\end{table}

\subsection{Human-Inspired evaluation method}
We developed three test papers using the SCI-CQA question bank to assess both multimodal models and a human evaluation team. The human team comprised nine graduate students, split into three groups of three. Each group completed one test, and the final score for each group was the average of the individual scores.
As shown in Tab.5, the human evaluation team consistently achieved scores around 90 points. In contrast, the highest model score was 41 points, attained by CogVLM on the data chart test. This significant gap highlights the substantial room for improvement in model performance.
We also measured inference speed, finding that models averaged 124.43 seconds per test on three L40 GPUs. In comparison, human evaluators required at least 2400 seconds to complete a test paper, demonstrating the efficiency advantage of automated chart understanding. Notably, human performance improved with additional time. In tests without time constraints, most human evaluators could correctly answer all multiple-choice and true/false questions, underscoring the value of time and thorough analysis in chart interpretation.

\subsection{Context-Based Complex Reasoning}
Fig.4 illustrates that providing contextual information or captions enhances the likelihood of models correctly answering previously unanswerable questions. 
In the data collection process, SCI-CQA included relevant contextual information and image captions. Using this data, we generated 50 unanswerable open-ended questions and evaluated model performance across five training iterations. As shown in Tab.6, models initially performed poorly without context or captions, scoring only 0.80 for data charts and 1.02 for flowcharts. Adding captions yielded modest improvements of 0.38 for data charts and 0.36 for flowcharts. However, the inclusion of contextual information led to substantial performance gains, with scores rising to 1.62 for data charts and 1.34 for flowcharts. This highlights the crucial role of contextual information in enhancing performance on complex reasoning tasks.
\begin{table}[h]
\centering
\renewcommand\arraystretch{1.2}
\resizebox{\columnwidth}{!}{
\begin{tabular}{c||ccc||c||c}
\hline\hline
No. & None      & Caption     & Context      & Data Open  & Flow Open   \\ \hline\hline
1   &\checkmark &             &              &  0.80    & 1.02      \\
2   &           &\checkmark   &              &  1.18    & 1.38      \\
3   &           &             &\checkmark    &  2.42    & 2.36      \\
4   &           &\checkmark   &\checkmark    &  2.48    & 3.18      \\
 \hline\hline
\end{tabular}}
\caption{Performance on unanswerable questions with additional context}
\label{table4}
\end{table}

However, comparing entries No. 3 and No. 4 shows that adding captions to contextual information yields only a modest improvement in performance. In contrast, the comparison between No. 3 and No. 1 indicates that for complex reasoning tasks, the most significant factor is the contextual information related to the image. Unfortunately, outside of SCI-CQA, few datasets offer chart-text pairs enriched with such detailed contextual information. SCI-CQA aims to address this gap by providing comprehensive data that advances chart understanding, particularly in ways that support practical scientific research applications.
\section{Conclusion}
This paper introduces SCI-CQA, a benchmark for chart understanding with a focus on scientific literature chart. It expands evaluation to include flowcharts and provides an expert-reviewed QA dataset for both data charts and flowcharts. To improve evaluation methods, the paper proposes an exam-inspired assessment approach and explores complex reasoning tasks using contextual information. A new data auto-annotation pipeline is also introduced. There are still some limitations. For instance, the data quality generated by the automated annotation pipeline remains suboptimal, even after multiple rounds of filtering. Additionally, manual review is costly, particularly for scientific charts, which require a higher level of education for reviewers compared to general image processing. As a result, the number of questions in our dataset still needs to be expanded. 

Future research could focus on the following directions: \textbf{Cross-subplot comparison and reasoning}: In multi-subplot scenarios, all models performed poorly in SCI-CQA, especially in questions involving cross-subplot comparison or reasoning. This remains a challenge to overcome. \textbf{Context-based complex reasoning with charts}: The ability to combine text and charts for complex reasoning, and even to provide scientific insights, represents the milestone of chart understanding. \textbf{Understanding complex visualizations in scientific literature:} These types of charts occupy a significant portion of scientific papers and are highly complex and diverse. How to effectively use models to automatically interpret these charts and further achieve comparison or reasoning is a topic worthy of in-depth study.

\section{Acknowledgment}
This work was supported by the 2035 Innovation Task - Scientific large model construction theory and method (Grants No.E4J10102).

\newpage
\bigskip
\bibliography{aaai25}
\section{Appendix}
\begin{table*}[h]
\centering
\renewcommand\arraystretch{1}
\resizebox{\textwidth}{!}{
\begin{tabular}{c | l | l l l l l}
\toprule
\toprule
Version & Source & Downloaded & Fetched & I-T pairs & Data\_HC & Flow\_HC \\ 
\midrule
V1 & CVPR, ICCV, ECCV & 8,155 & 9,593 & 43,710 & 3,570 & 2,525 \\
V2 & AAAI, IJCAI, KDD, ACM MM, SIGGRAPH, WWW, ICML, WACV, ACL & 33,360 & 50,477 & 92,800 & 15,861 & 2,788 \\
V3 & NIPS, CHI, ICLR & 13,042 & 15,466 & 66,250 & 11,826 & 1,037 \\
\midrule
\multirow{1}{*}{Total} & - & 54,557 & 75,506 & 202,760 & 31,257 & 6,350 \\
\bottomrule
\bottomrule
\end{tabular}}
\caption{Data source statistics}
\end{table*}

%
\subsection{Data source}
\begin{figure*}[t]
\centering
\includegraphics[width=0.7\textwidth]{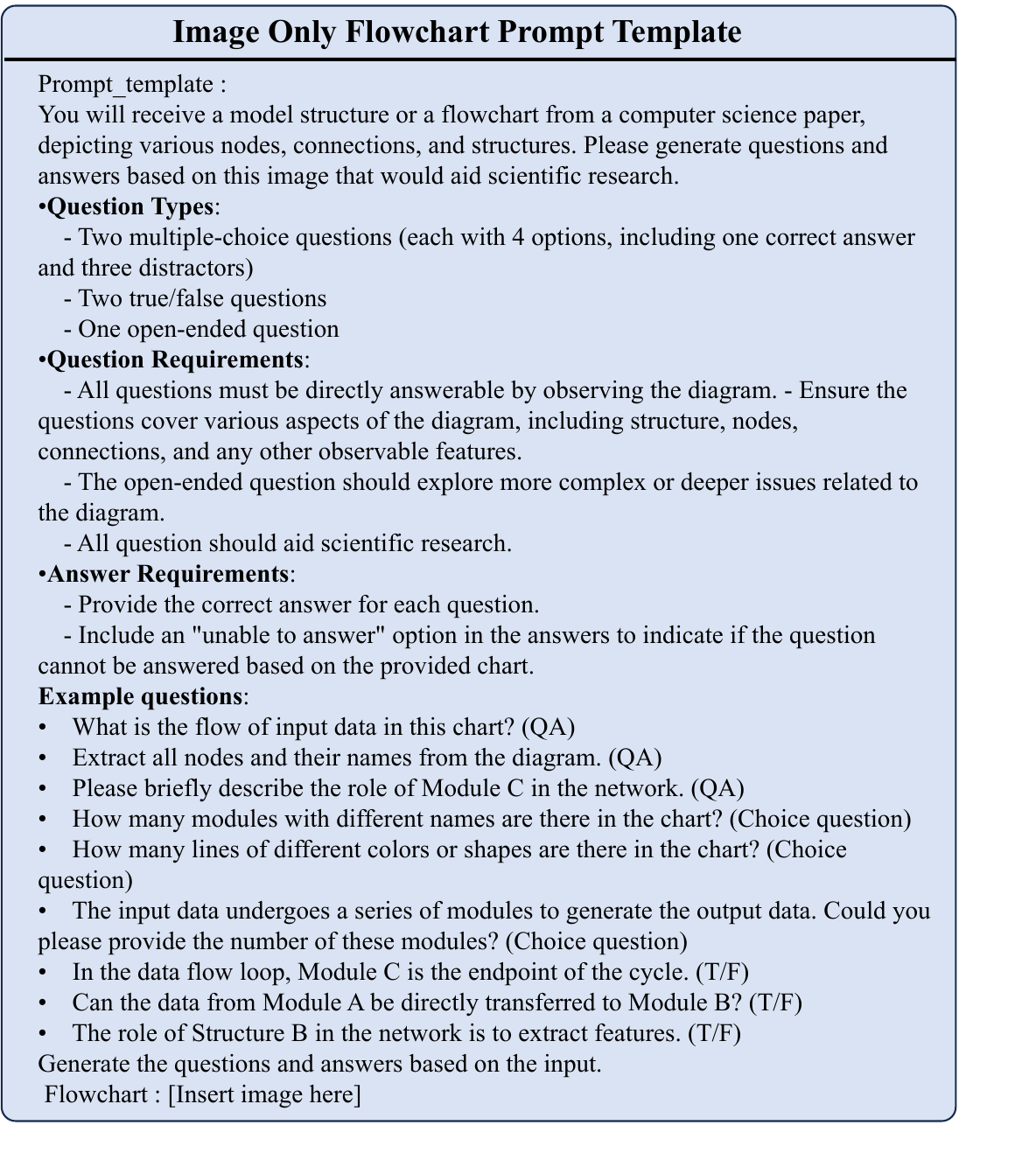} 
\caption{Generate question-and-answer pairs based solely on the provided flowchart image.}
\label{fig2}
\end{figure*}
\begin{figure*}[t]
\centering
\includegraphics[width=0.6\textwidth]{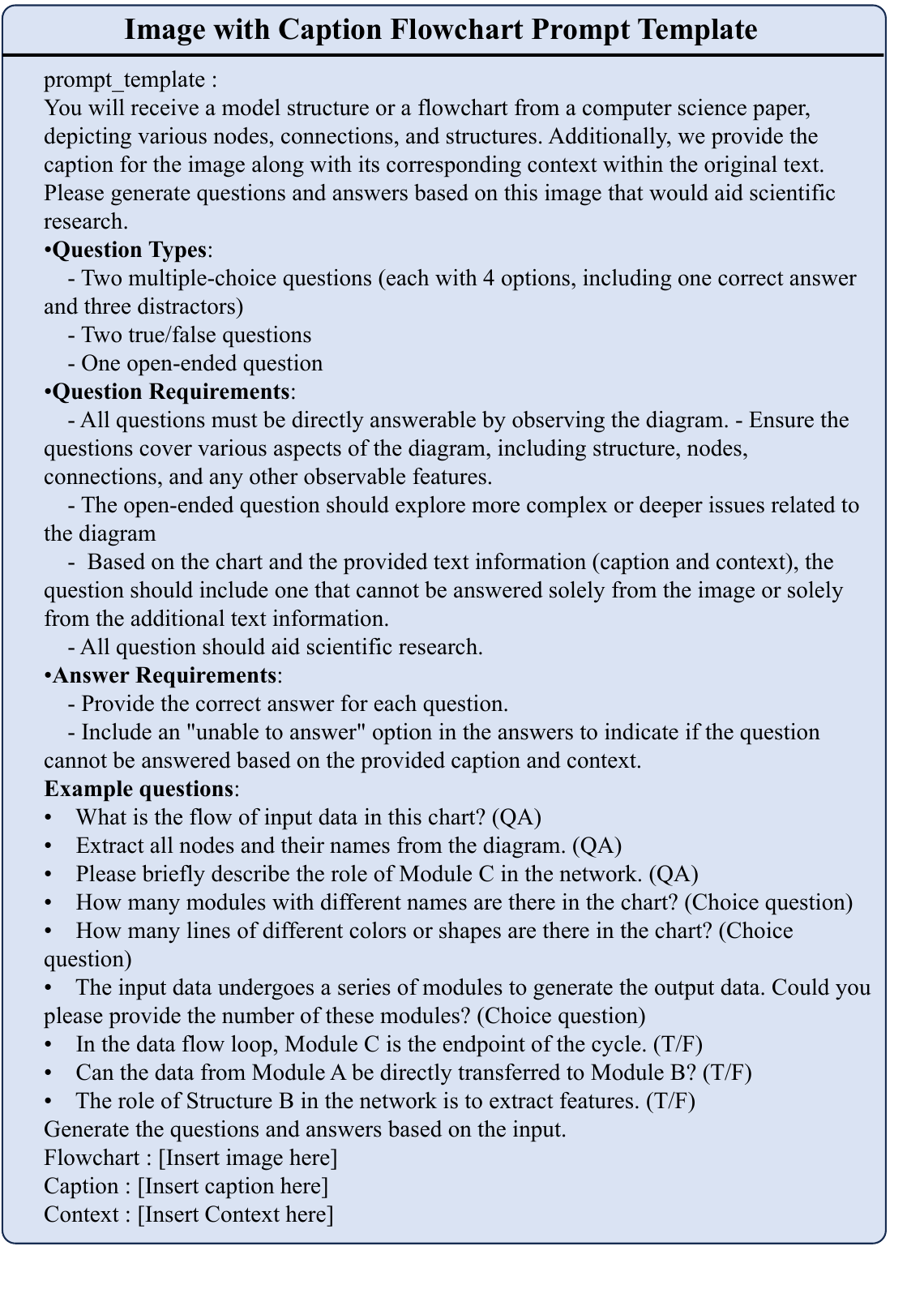} 
\caption{Generate question-and-answer pairs based on the provided flowchart image and additional text information.}
\label{fig2}
\end{figure*}
\begin{figure*}[t]
\centering
\includegraphics[width=0.59\textwidth]{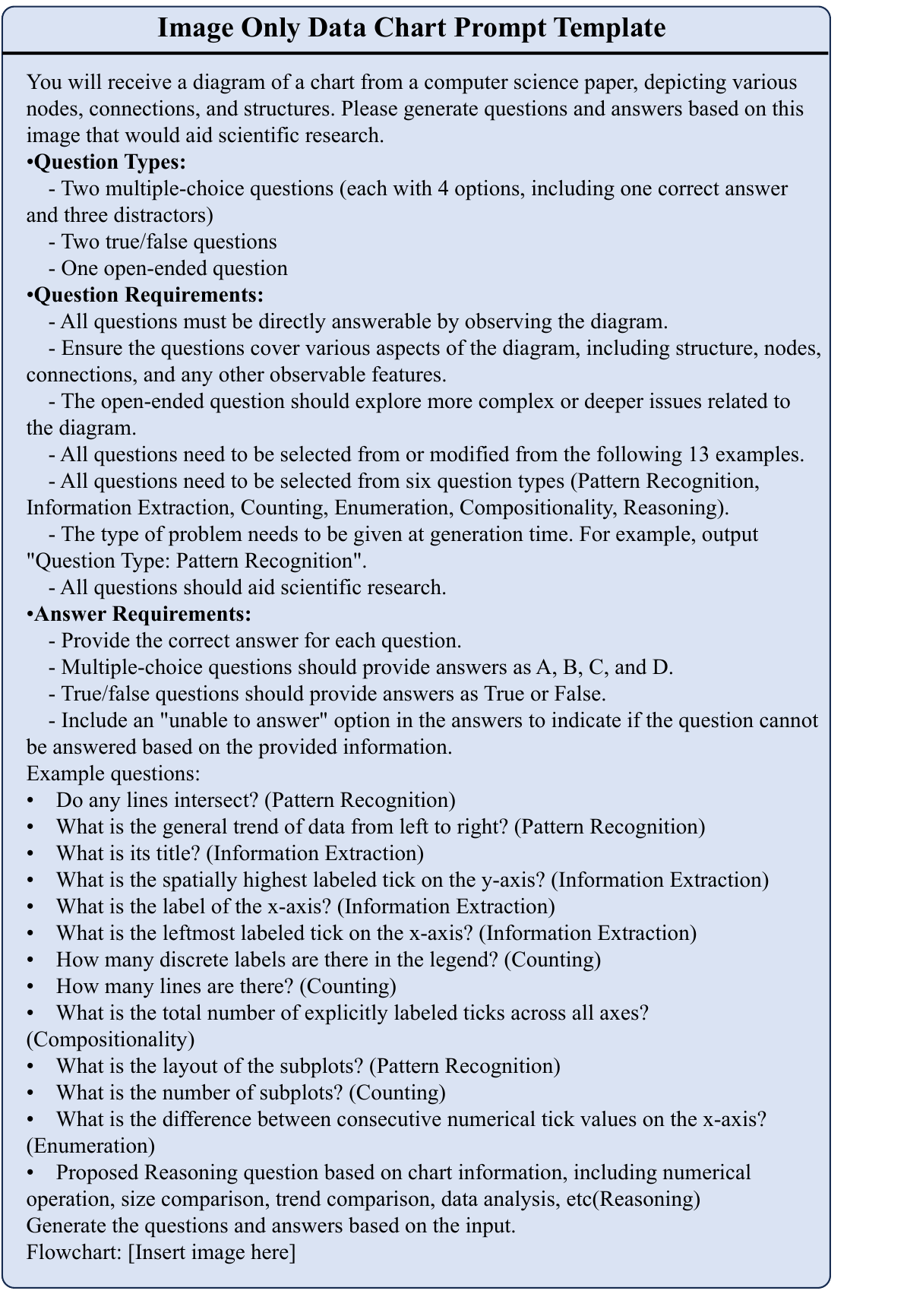} 
\caption{Generate question-and-answer pairs based solely on the provided data chart image.}
\label{fig2}
\end{figure*}
\begin{figure*}[t]
\centering
\includegraphics[width=0.7\textwidth]{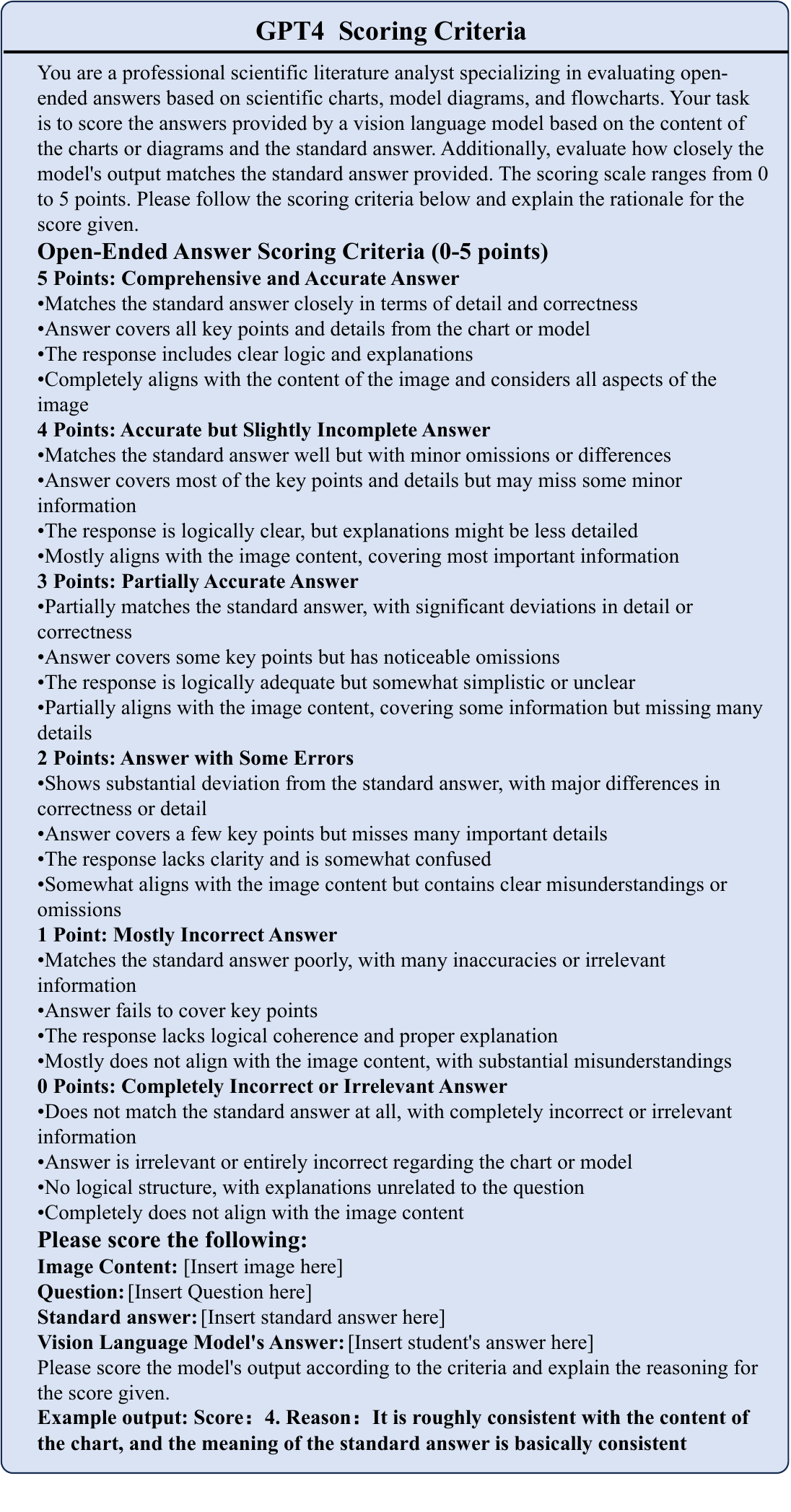} 
\caption{Criteria for Evaluating Open-Ended Answers by GPT-4o.}
\label{fig2}
\end{figure*}

The data is sourced from papers presented at 15 top-tier conferences in the field of computer science and artificial intelligence. As shown in Tab.7, our data is divided into three batches. The first batch initially consisted of 9,593 papers, from which 8,155 were successfully downloaded. The papers that were not downloaded were mainly due to the unavailability of their source files. From these papers, we extracted all the figures along with their related captions and context for subsequent data cleaning. After extraction, we obtained a total of 202,760 figure-caption pairs. Our task was not only to ensure that the charts had scientific value, but also to recognize that some illustrations, such as visualization diagrams, hold significant scientific value for understanding scientific literature. However, due to the complexity of the processing, we reserved these images for future use, focusing on flowcharts and data charts for now.

To ensure the quality of the extracted charts, these figure-caption pairs first underwent two rounds of model-based classification, followed by manual inspection. Due to the rigorous nature of this inspection, the number of figure-caption pairs we obtained is actually much lower than the true quantity available. However, this stringent strategy guarantees the quality of the figure-caption pairs in our dataset, which is crucial for creating a reliable benchmark. Future work can perform further cleaning on the original data we provide to obtain more data. For example, after cleaning, the Data\_CK dataset resulted in only 31,257 figure-caption pairs, representing just 15.4\% of all pairs. The ratio of flowcharts to data charts is approximately 1:5, which aligns with the characteristics of academic papers. Typically, a paper contains only 1-2 flowcharts (such as model structure diagrams or training process diagrams) but several data charts.

In addition to filtering images, we also filtered the text to ensure smoother language and to compress the token count, thereby reducing the cost of generating question-answer pairs later on. However, in subsequent experiments, we found that refining the context had little impact on the model's understanding. This might be because, despite the presence of some LaTeX symbols and language, the model—having been trained on a large corpus—exhibits strong robustness. GPT-4o is well-equipped to comprehend language that includes these special symbols.

\subsection{Q\&A Data generation}

We used GPT-4O as the model for data generation, specifically for creating initial question-and-answer pairs. In total, there are four different types of data generation methods. Firstly, based on different chart modalities, they can be categorized into flowcharts and data charts. Within these categories, they can further be divided into two types depending on whether additional text data is provided.\\

\noindent \textbf{[1].Generate QA pairs solely based on the flowchart.} 

For generating QA for flowcharts, we have established 9 questions that include multiple-choice, true/false, and open-ended questions. Using this approach, we generate 2 multiple-choice questions, 2 true/false questions, and 1 open-ended question for each image. Since flowcharts focus more on logic rather than distribution, our QA primarily targets aspects such as the function of modules, data flow, and the scientific nature of the structure. Additionally, we find that QA generation for flowcharts can better reflect complex reasoning or scientific assistance. For example, questions about a module's functionality can be expanded to compare the functions of two modules or explore how the design of the module contributes to performance improvements.

In addition, when only providing the image, the quality of the model's generated answers to \textbf{unanswerable questions} is significantly lower compared to when context is provided. Most of the generated questions focus on issues such as improvements to the modules present in the image. However, even with context provided, the model still struggles to answer these questions effectively.
\\

\noindent \textbf{[2].Generate QA pairs solely based on the data chart.} 

For generating questions about data charts, we referenced the excellent work by Charxiv. They defined questions into five domains: pattern recognition, enumeration, information extraction, counting, and composite questions. Based on these five categories, we introduced reasoning tasks as a sixth category. This task aims to deduce and summarize a reasoning task that aids scientific research based on the information in the chart. In all the generated question-answer pairs by GPT-4, reasoning tasks predominantly appear in open-ended questions. In the example QA, we used the 12 questions summarized in Charxiv and added examples of reasoning questions.\\

\noindent \textbf{[3].Generate QA pairs based on data/flowchart and text.} 

To address the above issue, we base our approach on using context to help the model generate questions that are unanswerable. This requires the model to not only propose questions that cannot be resolved with a single-modal input but also understand the complementary relationship between the two modalities. With the addition of information, questions that were previously unanswerable become solvable. 

\subsection{Case study}
In this section, we conducted case studies on the tested models. These studies help us better understand the shortcomings of the models in chart understanding.
\\

\noindent \textbf{[1].Multiple-Choice Questions} 

As shown in Figure 9, the question about the x-axis label tests the model's perception abilities. Due to the relatively small proportion of the label in the overall image, most open-source models provided incorrect answers. The question also assesses whether the model can accurately identify that the key information pertains to the second subplot. For example, Chartllama and DocOwlV1.5 incorrectly focused on the y-axis label of the first subplot, while Mini-Gemini identified the x-axis label of the first subplot. Additionally, Mini-Gemini's poor instruction-following capability resulted in its failure to provide options as per the example.

Figure 10 presents a multiple-choice question focused on a flowchart, which also serves as a perception task. Notably, all models failed to answer this question correctly. The complex logical relationships and visual elements made it difficult for the models to accurately identify the content in the upper right corner. Additionally, calculating the number of convolutional layers posed a significant challenge for the models. While it is straightforward for humans to count 12 layers from left to right, the models consistently provided answers with fewer than 12 layers.

Figure 11 is a scatter plot illustrating the distribution of different groups. The question pertains to whether the points are clustered or dispersed. As observed in the figure, the green points have the most widespread distribution.
\\

\noindent \textbf{[2].True / False Questions} 
\\
Figures 12-14 present true/false questions, with Figure 14 being particularly noteworthy. In this instance, the GPT series models unexpectedly provided incorrect answers, whereas some open-source models, like LLaVA and Mini-Gemini, correctly answered the question. This demonstrates that our questions effectively differentiate model performance and highlights the increased difficulty of correctly answering questions related to flowcharts. These challenges arise because models must not only interpret the visual elements but also comprehend the textual content, such as understanding concepts like loops and feedback mechanisms.
\\

\noindent \textbf{[3].Open-Ended Questions} 
\\
Figures 15 and 16 highlight open-ended questions, with scores assigned by GPT-4o according to established evaluation criteria. The question in Figure 15 involves comparing two models across two different charts, posing a significant challenge. Typically, chart understanding tasks focus on reasoning or perception within a single chart or a specific subplot. However, some questions in SCI-CQA require comparing multiple metrics across different charts. 

In Figure 15, although InternVL2 correctly identified the overall trend, it inaccurately reported the detailed numerical growth and only focused on the chart on the left, resulting in a score of just 1 point. On the other hand, while LLaVA-v1.6 accurately captured the general trend and partially correct analysis, it failed to provide precise numerical growth, reflecting the complexity of the task and the difficulty models face in accurately processing such detailed comparisons.

In Figure 16, a seemingly straightforward data flow question posed a significant challenge for the models. Although most of the perception tasks related to the question were correctly handled by the models, both struggled when it came to reasoning about the data flow. LLaVA, in particular, exhibited severe hallucinations, providing a response that did not align with the question's intent. Similarly, InternVL2 incorrectly included the top-left module as part of the data flow, further highlighting the difficulty these models face in accurately reasoning about the flow of data, despite successfully completing related perception tasks.

We found a significant disparity between open-source and proprietary models in open-ended questions. This gap may arise because open-ended questions often involve conceptual topics such as loops and convolutions. Due to dataset differences, some open-source models struggle to grasp these concepts effectively, leading to more frequent hallucinations in their responses.
\\

\noindent \textbf{[4].Conclusion}
Through case studies, we observed that high-resolution visual modules significantly enhance performance on perception tasks. For instance, InternVL2 and Qwen-VL, which retrained their visual encoders and increased input image resolution, demonstrated improved perceptual capabilities. In contrast, Chartllama, with its lower resolution, struggled with the perception tasks in Figure 10.

Moreover, a model’s reasoning ability is closely related to its training data. InternVL2, with its extensive pre-training dataset including chart data, showed noticeable improvements and even matched proprietary models in some tasks. This suggests that GPT-series models, with their more comprehensive training, including large amounts of scientific literature, exhibit strong performance in both perceptual and reasoning tasks.

\begin{figure*}[t]
\centering
\includegraphics[width=0.8\textwidth]{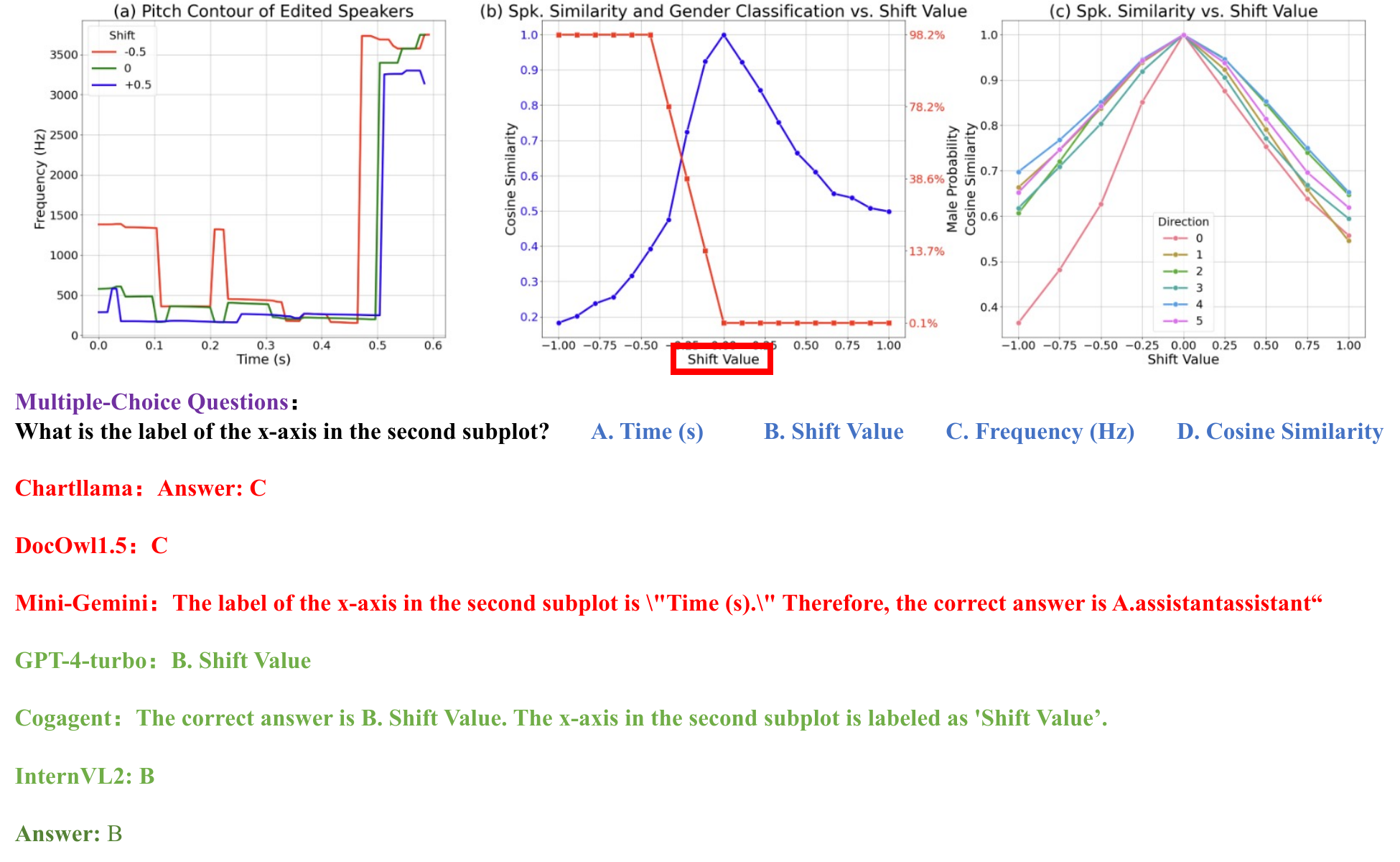} 
\caption{The multiple-choice question regarding the data chart. The labels on the x-axis highlighted with a red box.}
\label{fig5}
\end{figure*}

\begin{figure*}[t]
\centering
\includegraphics[width=0.8\textwidth]{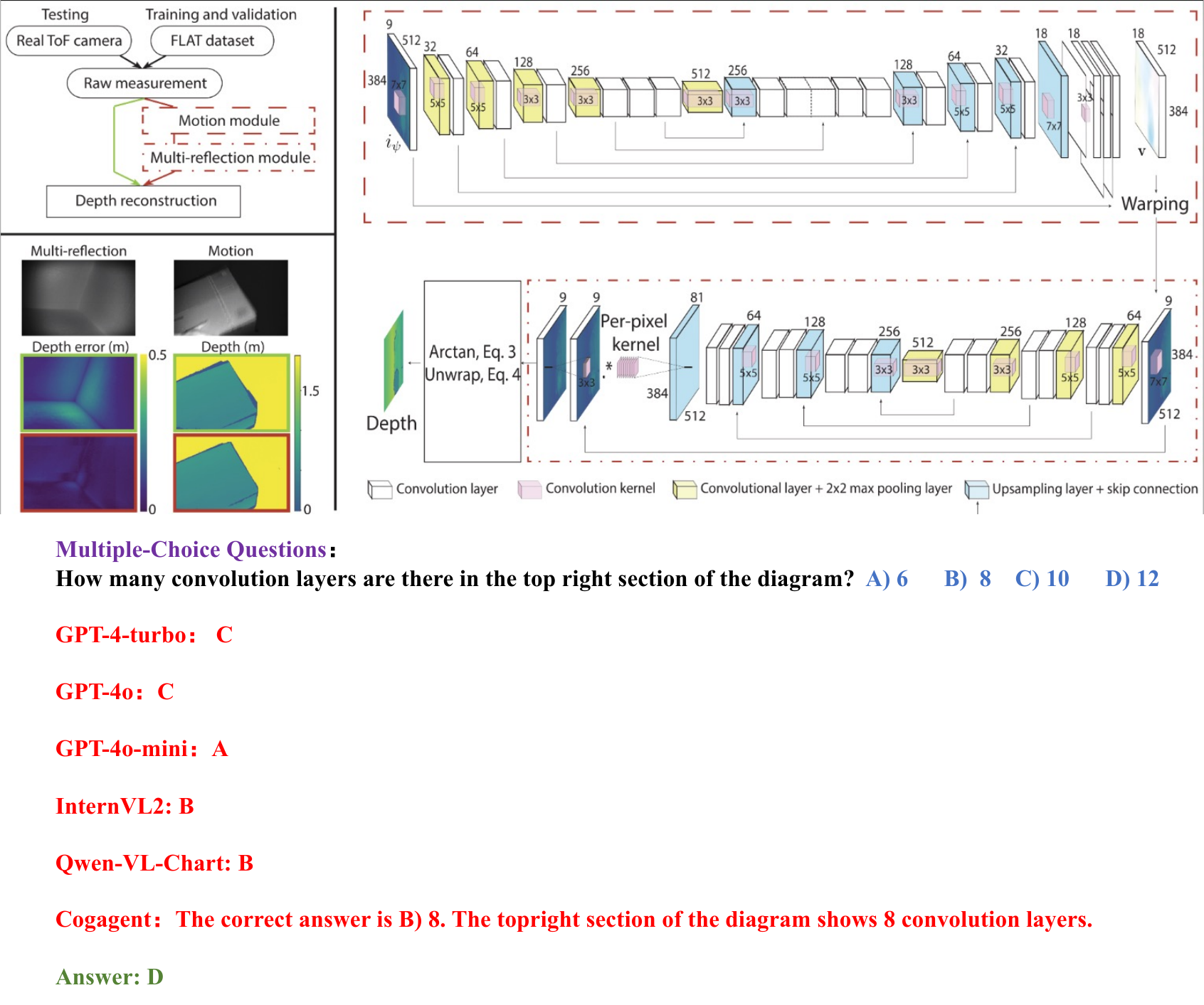} 
\caption{The multiple-choice question regarding the flowchart. The number of convolutional layers in the top right corner is 12 layers.}
\label{fig6}
\end{figure*}

\begin{figure*}[t]
\centering
\includegraphics[width=0.9\textwidth]{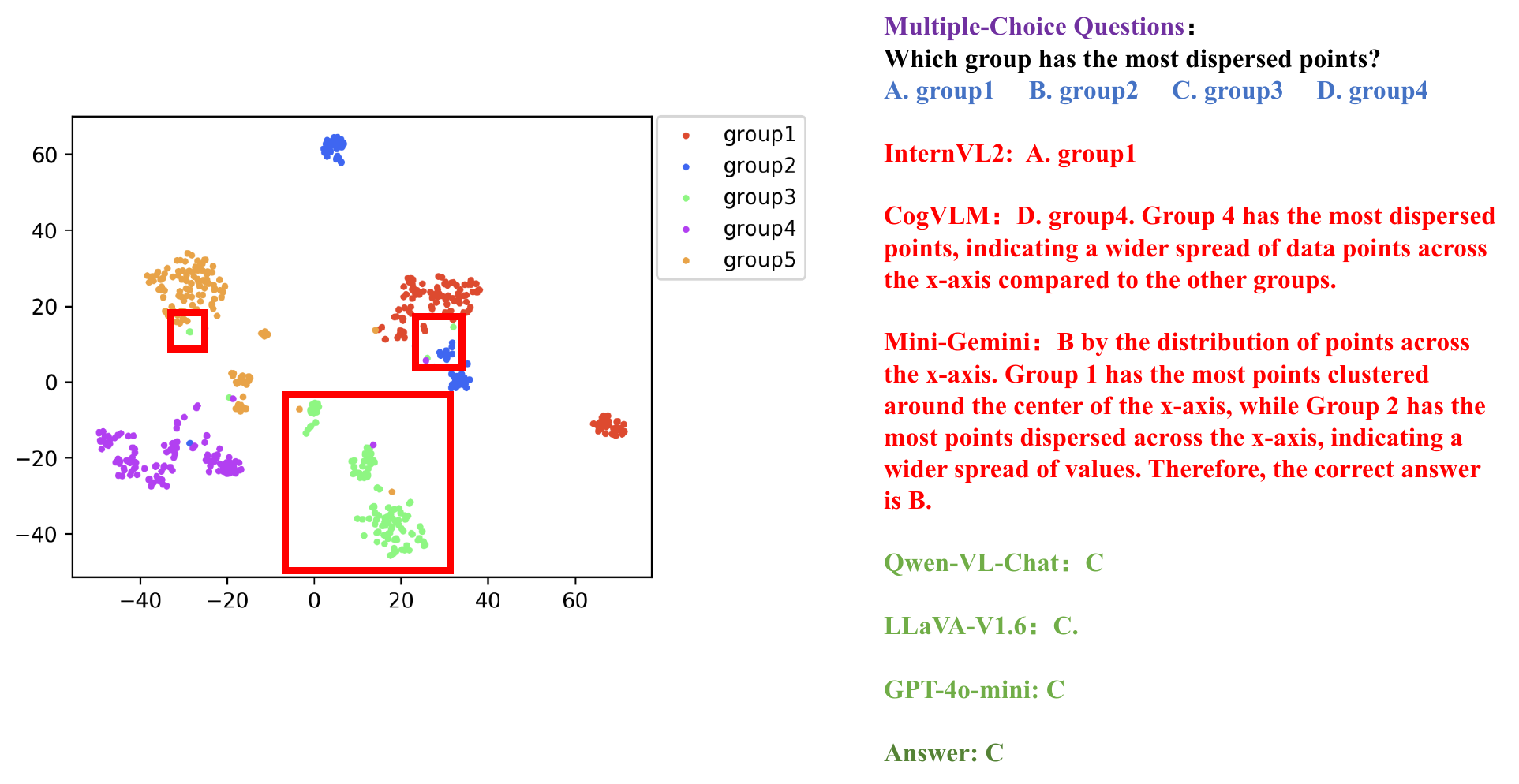} 
\caption{The multiple-choice question regarding the data chart. The red box in the figure clearly shows that the distribution of green dots is the most scattered.}
\label{fig7}
\end{figure*}

\begin{figure*}[t]
\centering
\includegraphics[width=0.9\textwidth]{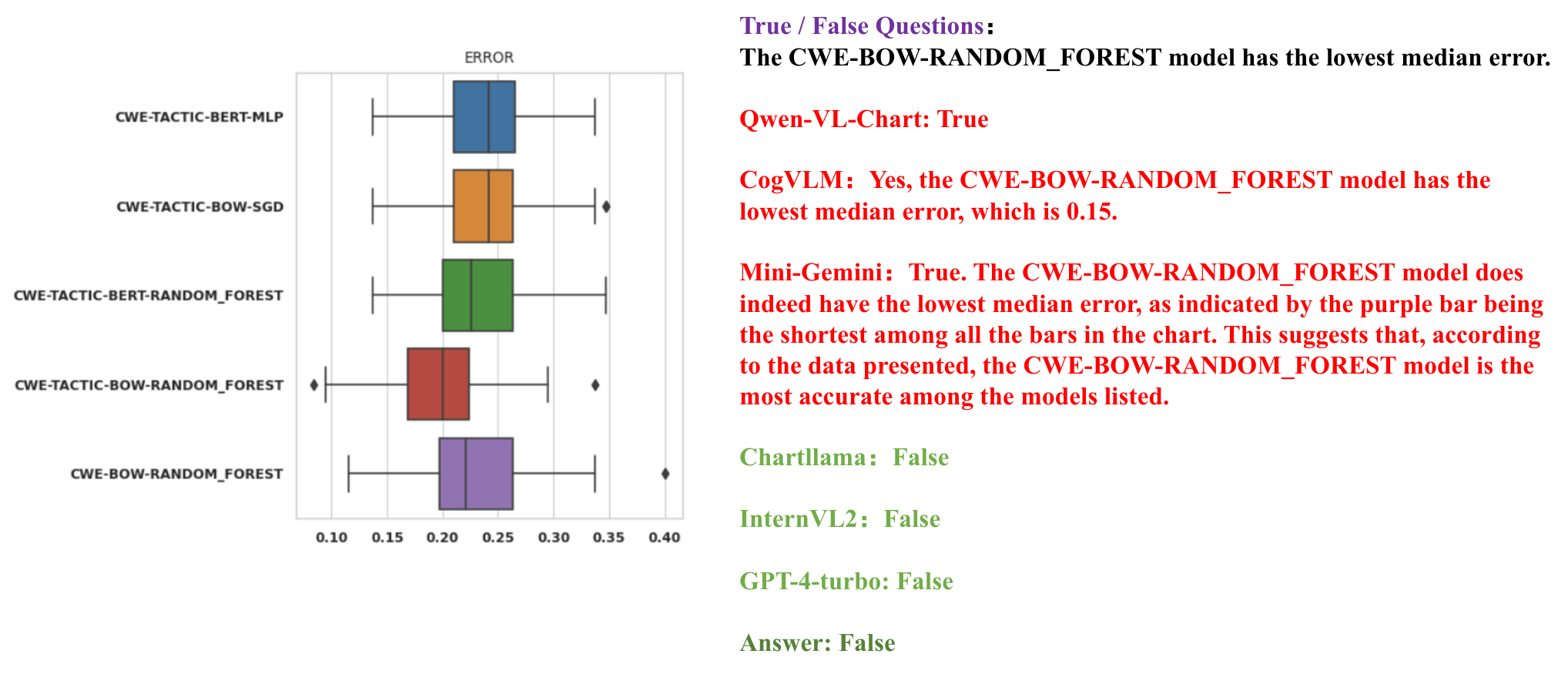} 
\caption{The True/False question regarding the data chart. The figure shows that the median error of CWE-BOW-RANDOM\_FOREST is not the lowest.}
\label{fig8}
\end{figure*}

\begin{figure*}[t]
\centering
\includegraphics[width=0.9\textwidth]{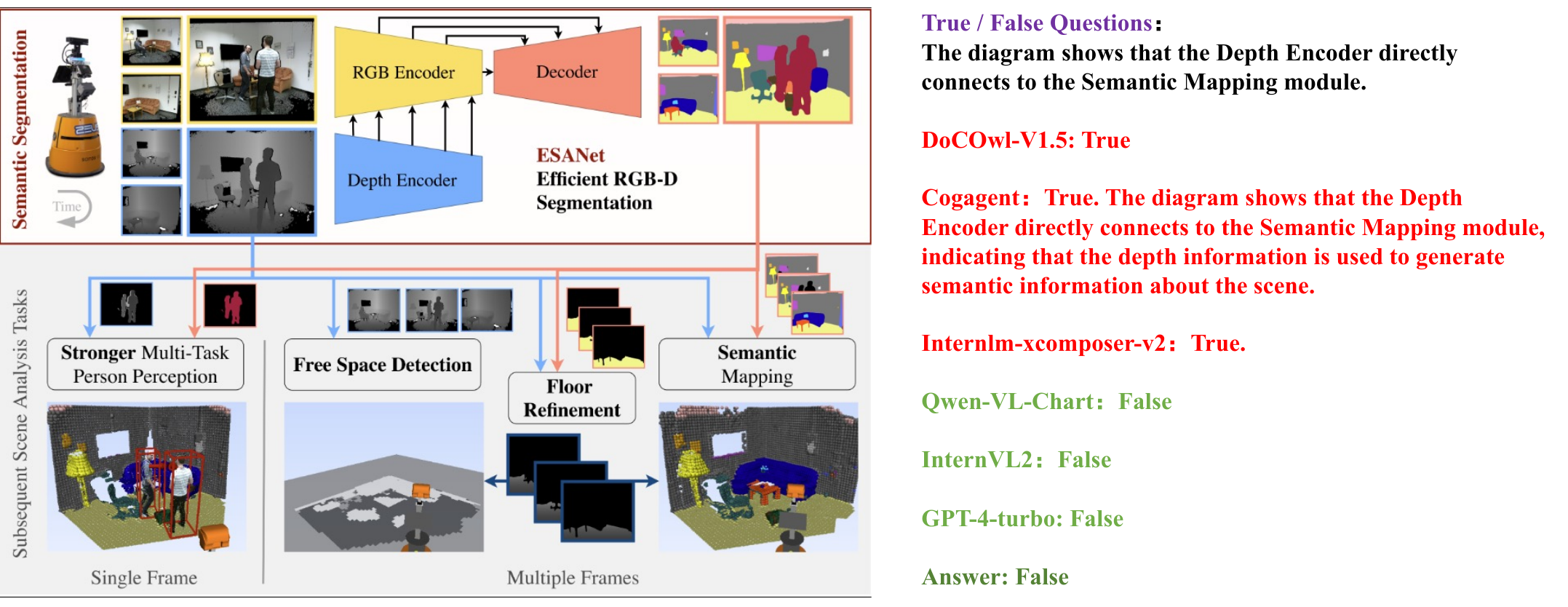} 
\caption{The True/False question regarding the flowchart. The Depth Encoder is not directly
connects to the Semantic Mapping module.}
\label{fig9}
\end{figure*}

\begin{figure*}[t]
\centering
\includegraphics[width=0.9\textwidth]{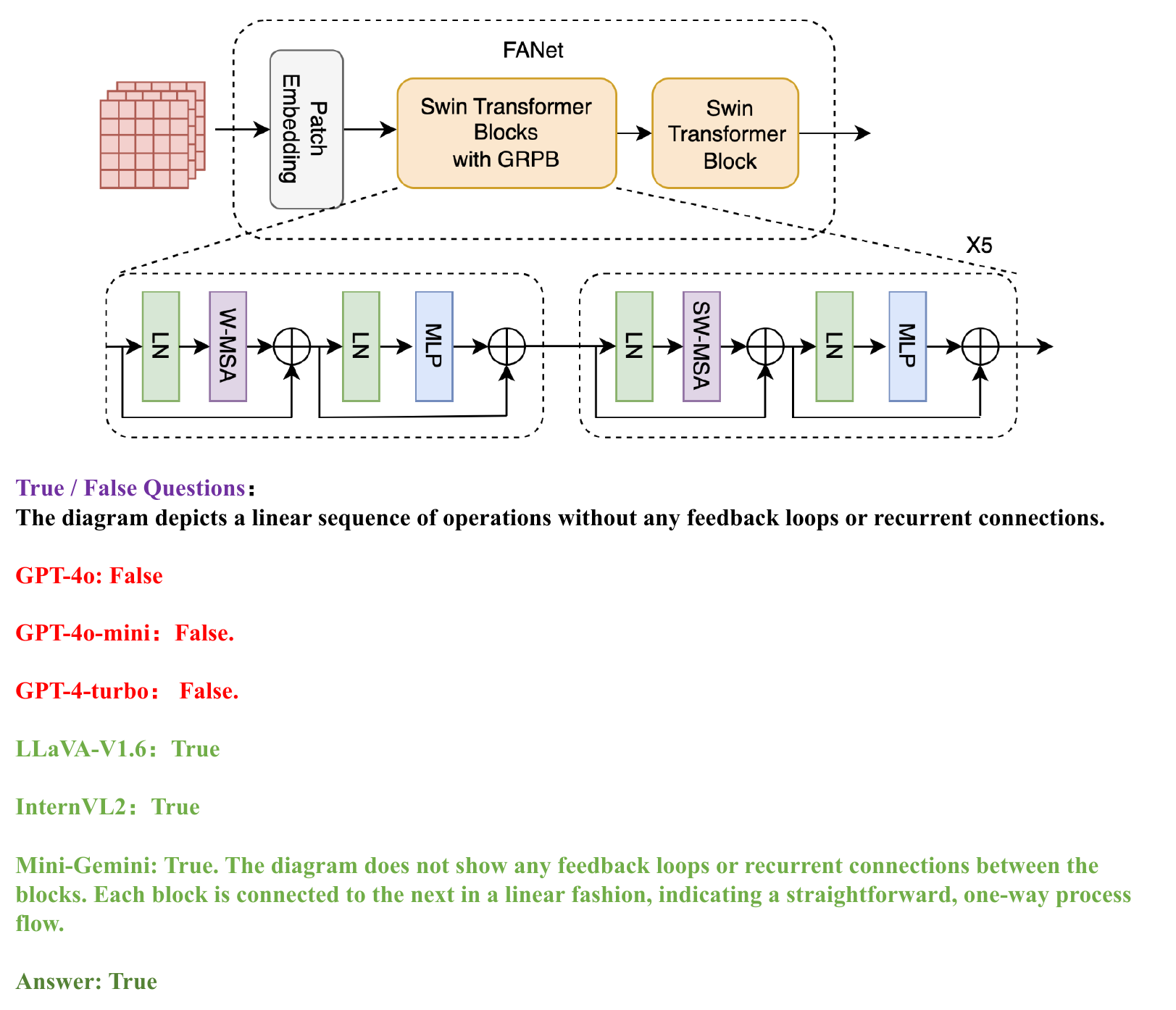} 
\caption{The True/False question regarding the flowchart.}
\label{fig10}
\end{figure*}

\begin{figure*}[t]
\centering
\includegraphics[width=1\textwidth]{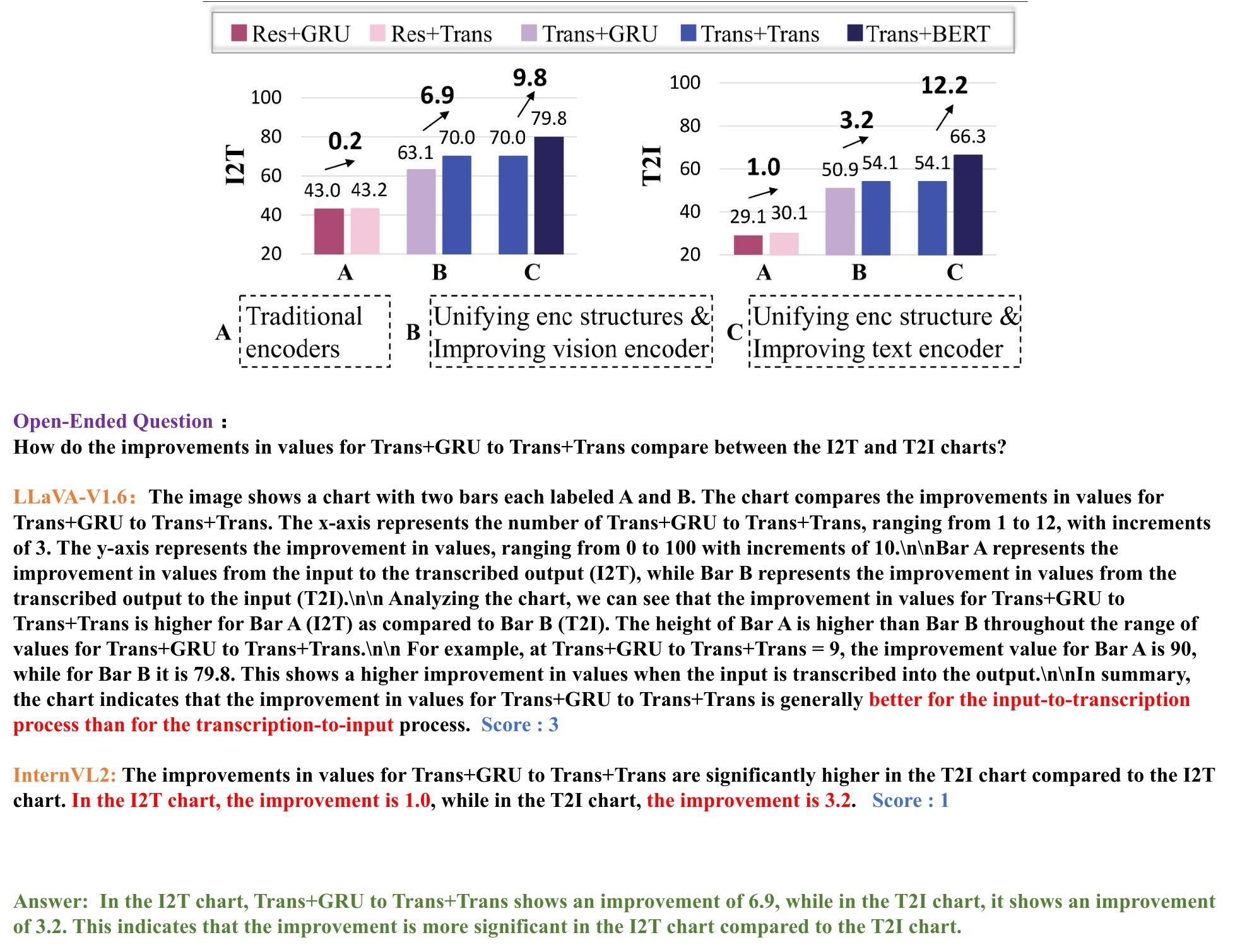} 
\caption{The Open-Ended question regarding the flowchart.}
\label{fig11}
\end{figure*}

\begin{figure*}[t]
\centering
\includegraphics[width=1\textwidth]{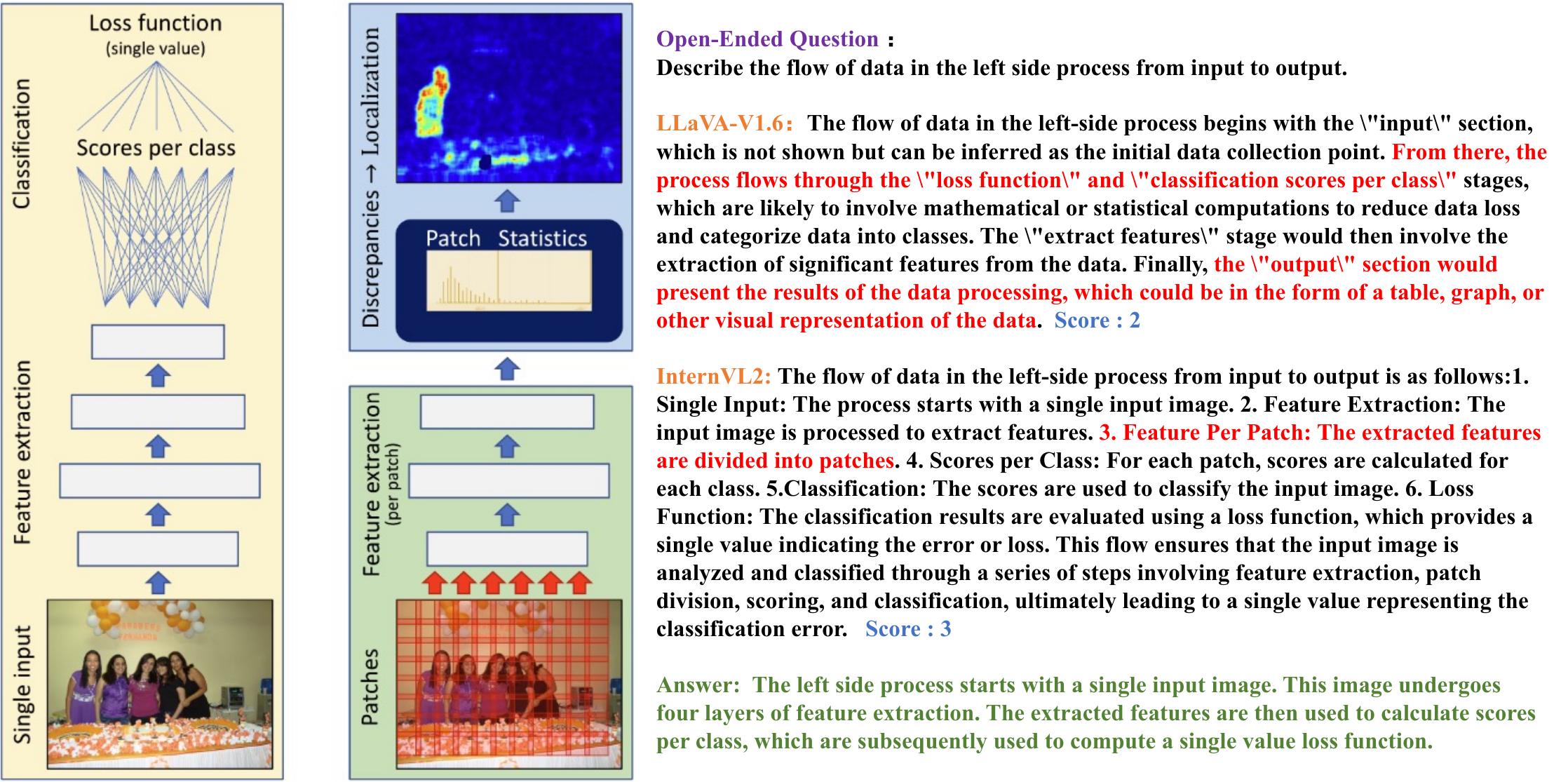} 
\caption{The Open-Ended question regarding the flowchart.}
\label{fig5}
\end{figure*}
\end{document}